\pdfoutput=1
\documentclass[10pt, logo, onecolumn, copyright]{nvidiatechreport}

\usepackage{mdframed}
\usepackage[utf8]{inputenc} %
\usepackage[T1]{fontenc}    %

\usepackage{amsfonts}       %
\usepackage{nicefrac}       %
\usepackage{microtype}      %
\usepackage[dvipsnames]{xcolor}         %
\usepackage{multirow}
\usepackage{multicol}
\usepackage{graphicx}
\usepackage[numbers]{natbib}
\usepackage{tabto}
\usepackage{xspace}
\usepackage{amsmath}
\usepackage{adjustbox}
\usepackage{enumitem}
\usepackage{subcaption}
\usepackage{wrapfig}
\usepackage{longtable}
\usepackage{dblfloatfix}

\usepackage{footmisc}

\usepackage{float}

\usepackage{natbib}

\usepackage{hyperref}
\usepackage{url}
\usepackage{booktabs}
\usepackage{multirow}
\usepackage{enumitem}
\usepackage{adjustbox}
\usepackage{tabularx}
\usepackage{arydshln}
\usepackage{wrapfig}

\usepackage{multirow}
\usepackage{colortbl}
\usepackage{amsmath}
\usepackage{makecell}
\usepackage{hhline}
\usepackage{array}
\usepackage{diagbox}
\usepackage{adjustbox}

\usepackage{fp} %

\usepackage{xcolor}
\usepackage[most]{tcolorbox}
\usepackage{lipsum}

\usepackage{natbib}
\usepackage{makecell,hhline}
\usepackage{booktabs}
\usepackage{mdframed}
\usepackage[utf8]{inputenc} %
\usepackage[T1]{fontenc}    %

\usepackage{amsfonts}       %
\usepackage{nicefrac}       %
\usepackage{microtype}      %
\usepackage[dvipsnames]{xcolor}         %
\usepackage{multirow}
\usepackage{multicol}
\usepackage{graphicx}
\usepackage{xspace}
\usepackage{amsmath}
\usepackage{adjustbox}
\usepackage{enumitem}
\usepackage{subcaption}
\usepackage{wrapfig}
\usepackage{longtable}
\usepackage{dblfloatfix}
\usepackage[table]{xcolor}

\usepackage{fp} %
\usepackage{array,xspace} %

\usepackage{xcolor}
\usepackage[most]{tcolorbox}
\usepackage{lipsum}

\newcommand{\app}{\raise.17ex\hbox{$\scriptstyle\sim$}}

\newcolumntype{x}[1]{>{\centering\arraybackslash}p{#1pt}}
\newcolumntype{y}[1]{>{\raggedright\arraybackslash}p{#1pt}}

\newlength\savewidth
\newcommand{\tablestyle}[2]{\setlength{\tabcolsep}{#1}\renewcommand{\arraystretch}{#2}\centering\footnotesize}
\makeatletter\renewcommand\paragraph{\@startsection{paragraph}{4}{\z@}
  {.5em \@plus1ex \@minus.2ex}{-.5em}{\normalfont\normalsize\bfseries}}\makeatother

\newcommand{\modelname}{OmniVinci\xspace}

\newcommand{\alignmodelname}{OmniAlignNet\xspace}

\definecolor{nvgreen}{RGB}{133,184,55}
\definecolor{lightgreen}{RGB}{230,240,236}
\definecolor{deepgreen}{HTML}{195B24}

\usepackage{tikz}

\usepackage{pgfplots}
\pgfplotsset{compat=1.18} %

\newcounter{keyinsight}

\tcbset{
  highlightstyle/.style={
    colback=white,
    colframe=nvgreen,
    arc=3mm,        
    boxrule=0.8pt, 
    left=5pt, right=5pt, top=5pt, bottom=5pt,
    enhanced
  }
}

\newcommand{\keyinsight}[1]{%
  \refstepcounter{keyinsight}%
    \textbf{Key Insight \thekeyinsight.} #1
}

\usepackage{titletoc}

\title{\modelname: Enhancing Architecture and Data for Omni-Modal Understanding LLM}

\author{
Hanrong Ye$^{\dagger*}$ ~
Chao-Han Huck Yang$^*$ ~
Arushi Goel$^*$ ~
Wei Huang$^*$ ~
Ligeng Zhu$^*$ ~
Yuanhang Su$^*$ ~
Sean Lin$^*$ ~
An-Chieh Cheng$^*$ ~
Zhen Wan$^*$ ~
Jinchuan Tian$^*$ ~
Yuming Lou$^*$ ~
Dong Yang$^*$ ~
Zhijian Liu ~
Yukang Chen ~
Ambrish Dantrey ~
Ehsan Jahangiri ~
Sreyan Ghosh ~
Daguang Xu ~
Ehsan Hosseini-Asl ~
Danial Mohseni Taheri ~
Vidya Murali ~
Sifei Liu ~
Yao Lu ~
Oluwatobi Olabiyi ~
Yu-Chiang Frank Wang ~
Rafael Valle ~
Bryan Catanzaro ~
Andrew Tao ~
Song Han ~
Jan Kautz ~
Hongxu Yin$^{\S\dagger*}$ ~
Pavlo Molchanov$^{\S}$ \\
\vspace{2mm}
{\normalsize NVIDIA} \\
\vspace{2mm}
{\normalsize
 $^*$Core Contribution ~~ $^\dagger$Corresponding Authors ~~ $^{\S}$Equal Advisory \\
\vspace{2mm}
\normalsize \href{https://github.com/NVlabs/OmniVinci}{Code} ~~~~
\normalsize \href{https://huggingface.co/nvidia/omnivinci}{Model} ~~~~
\normalsize \href{https://nvlabs.github.io/OmniVinci}{Webpage}
}
}

\begin{abstract}
\textbf{Abstract}\\
Advancing machine intelligence requires developing the ability to perceive across multiple modalities, much as humans sense the world.
We introduce \modelname, an initiative to build a strong, open-source, omni-modal LLM.
We carefully study the design choices across model architecture and data curation.
For model architecture, we present three key innovations:
(i) \alignmodelname for strengthening alignment between vision and audio embeddings in a shared omni-modal latent space;
(ii) Temporal Embedding Grouping for capturing relative temporal alignment between vision and audio signals; and
(iii) Constrained Rotary Time Embedding for encoding absolute temporal information in omni-modal embeddings. 
We introduce a curation and synthesis pipeline that generates 24M single-modal and omni-modal conversations. We find that modalities reinforce one another in both perception and reasoning. Our model, \modelname, outperforms Qwen2.5-Omni with +19.05 on DailyOmni (cross-modal understanding), +1.7 on MMAR (audio), and +3.9 on Video-MME (vision), while using just 0.2T training tokens - a 6$\times$ reduction compared to Qwen2.5-Omni’s 1.2T.
We finally demonstrate omni-modal advantages in downstream applications spanning robotics, medical AI, and smart factory.

\end{abstract}

\begin{document}
\maketitle

\begin{figure*}[h]
\centering
\includegraphics[width=1\linewidth]{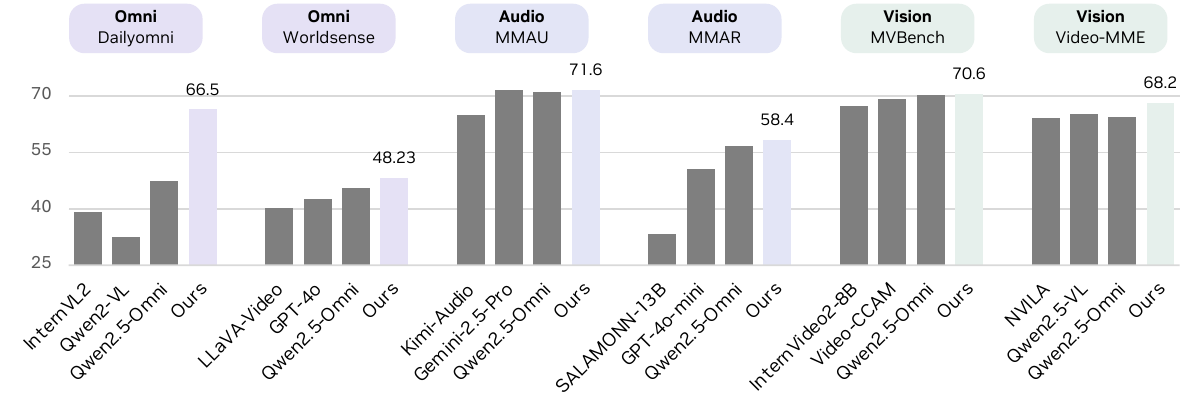}
\vspace{-20pt}
\caption{\modelname demonstrates strong performance across widely used omni-modal (+19.05 on Dailyomni), audio (+1.7 on MMAR), and vision (+3.9 on Video-MME) understanding benchmarks. 
}
\label{fig:teaser}
\vspace{-10pt}
\end{figure*}

\section{Introduction}

The progress of multimodal LLMs has demonstrated appealing applications when LLMs learn to see with vision~\citep{lin2023vila,liu2023llava,alayrac2022flamingo} or listen with audio~\citep{goel2025audio,chu2023qwen,tang2023salmonn}. Recent work has enabled joint video-audio alignment, further unifying their strengths towards general intelligence~\citep{openai2024gpt4o,wu2024next,tang2024codi,ye2024x,abouelenin2025phi,xu2025qwen25omni}. However, training such an omni-modal system can be expensive and challenging across many dimensions, as it relies on proper choices of network architecture and data recipe.

This work presents a systematic exploration of developing omni-modal LLMs aiming to enable simultaneous understanding of vision, audio (encompassing both natural sounds and human speech), and language. 
We ablate and validate the design choices overseeing model architecture design, data curation, and training strategy. 
For model architecture, we introduce a new framework to harmonize vision and audio embeddings in a unified omni-modal embedding space, featuring three new techniques: (i) \textit{\textit{\alignmodelname}} that learns to construct a modality-shared space to align vision and audio embeddings from the same video; \text{(ii)} \textit{\textit{Temporal Embedding Grouping}} that divides the time dimension into multiple chunks and reorganizes the vision and audio embeddings according to their timestamps to align with the corresponding chunks; (iii) \textit{\textit{Constrained Rotary Time Embedding}} to directly insert periodic temporal information into vision-audio embeddings. 
We observe noticeable performance improvements with these techniques, as shown later in our experiments. 
On the data front, we curate 24 million high-quality multimodal conversation samples that span a diverse set of tasks across audio, video, and image domains, including both modal-specific conversations and omni-modal conversations.
We tackle the scarcity of omni-modal data by exploiting existing video-with-audio QA data, which implicitly encodes omni-modal signals (\textit{implicit learning}). To further facilitate omni-modal learning, we generate synthetic conversations with explicit omni-modal labels (\textit{explicit learning}).

Our findings enable a frontier omni-modal model, named \modelname. 
See a quick performance comparison in Figure~\ref{fig:teaser} and more in our experimental section. Compared to prior art such as Qwen2.5-Omni and Gemini-2.5-Pro, \modelname further pushes the boundary of various multimodal understanding tasks, with gains of +2.83\% on WorldSense and +19.05\% on Dailyomni for joint vision-audio understanding, +1.7\% on MMAR for audio understanding, and +3.9\% on Video-MME for vision understanding. \modelname also pushes on efficiency fronts using only 0.2T training tokens, around 6$\times$ fewer than Qwen2.5-Omni's 1.2T tokens. 
More encouragingly, we observe the synergy between audio and video, not only for perception, but also for reasoning. Finally, we demonstrate that \modelname has enabled or improved a wide range of important downstream applications, including robotics, video broadcasting, medical, and smart factory use cases.

\section{Model Architecture}
\label{sec:model_arch}

\begin{figure*}[t]
\centering
% \vspace{-40pt}
\includegraphics[width=\linewidth]{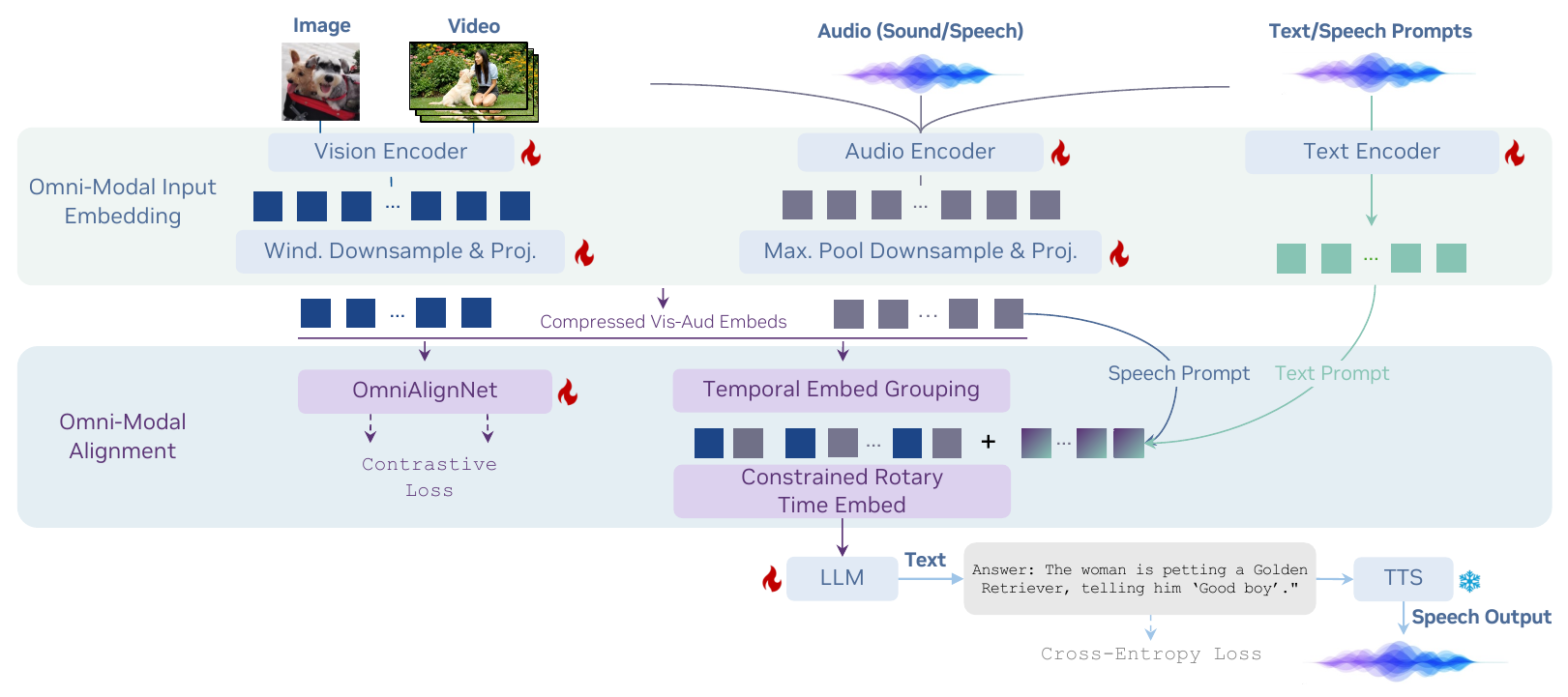}
\vspace{-10pt}
\caption{We introduce a foundation model for omni-modal understanding. Our model blends information from vision, audio, and text modalities into a unified omni-modal token sequence via the proposed omni-modal alignment mechanism. 
}
\vspace{-10pt}
\label{fig:arch}
\end{figure*}

The key objective of model architecture design is to support composable cross-modal understanding through integrating heterogeneous input from images, videos, audio, and text, into a shared omni-modal latent space. 
As shown in Figure~\ref{fig:arch}, we adopt an auto-regressive regime to encode visual and audio signals, and then align them as input of LLM backbone.

\noindent\textbf{Omni-Modal Input Embedding.}
To simplify the network design, 
we (i) decompose video into a sequence of temporally correlated images and audio, and (ii) employ a unified audio encoder to handle both acoustic and speech information in context and prompt. We present the encoder sharing paths in Figure~\ref{fig:arch}, and describe the details of encoding streams in Appendix~\ref{sec:omni_embed}.

\subsection{Omni-Modal Alignment Mechanism}
\label{sec:omni_align_mechanism}

We next integrate embeddings from all modalities into a unified latent space as input for LLM.

\begin{wrapfigure}{r}{0.5\textwidth} %
  \centering
  \vspace{-\baselineskip}              %
  \includegraphics[width=\linewidth]{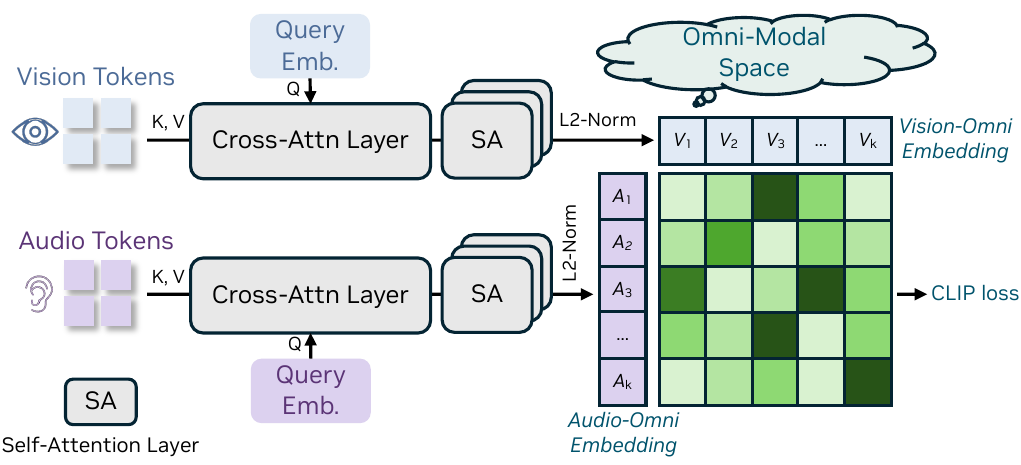} %
  \vspace{-10pt}
  \caption{Illustration of the proposed \alignmodelname module. 
  \vspace{-5pt}
  }
  \label{fig:imagine_module}
\end{wrapfigure}

\noindent\textbf{\alignmodelname module.}
For a given input video, the audio and vision streams have an inherent semantic connection, providing complementary information for each other. Such a correlation provides a natural way to more effectively learn and align vision and audio embeddings in the unified latent space. 
To this end, we propose  \alignmodelname, which strengthens the learning of vision and audio embeddings via exploiting their complementary information. As illustrated in Figure~\ref{fig:imagine_module}, the \alignmodelname module first maps visual and audio embedding sequences (outputs of modality-specific projectors) into a shared latent embedding space and then aligns them via contrastive learning, inspired by ImageBind~\citep{girdhar2023imagebind}.

Given an input video with an accompanying audio stream, we denote the sequence of visual embeddings produced by the visual projection layer as $\mathbf{E}_{v} \in \mathbb{R}^{N_{v} \times C}$ and the sequence of audio embeddings produced by the audio projection layer as $\mathbf{E}_{a} \in \mathbb{R}^{N_{a} \times C}$, with $N_{v}$ and $N_{a}$ represent the number of visual and audio embeddings, respectively, while $C$ denotes the latent dimensionality. To align representations, we initialize a vision query embedding $\mathbf{Q}_{v} \in \mathbb{R}^{1 \times C}$ and an audio query embedding $\mathbf{Q}_{a} \in \mathbb{R}^{1 \times C}$. These queries are used to project $\mathbf{E}_{v}$ and $\mathbf{E}_{a}$ into fixed-size embeddings of shape $(1 \times C)$. 
Suppose each batch has $K$ videos, the projected features are then processed through three layers of self-attention modules and L2 normalized, yielding the vision-omni embedding $\mathbf{V} \in \mathbb{R}^{K\times C}$ and the audio-omni embedding $\mathbf{A} \in \mathbb{R}^{K\times C}$, respectively, in a modality-shared latent space.

With embeddings $\mathbf{V}$ and  $\mathbf{A}$ in the shared latent space, we now apply CLIP-style contrastive loss~\citep{radford2021clip} on the output embeddings to minimize intra-sample cross-modal distance, while maximizing inter-sample cross-modal distance. 
Let $\{\mathbf{V}_i, \mathbf{A}_i\}_{i=1}^{K}$ be the set of L2-normalized visual and audio embeddings for a batch of $K$ video clips. The similarity between the $i$-th visual embedding and the $j$-th audio embedding is computed as their dot product, $s_{ij} = \mathbf{V}_i^T \mathbf{A}_j$. The contrastive loss is then formulated as a symmetric cross-entropy loss over the similarity score. The loss for aligning vision to audio ($L_{v \to a}$) and audio to vision ($L_{a \to v}$) is:
\begin{equation}
\begin{aligned}
    L_{v \to a} &= -\frac{1}{N} \sum_{i=1}^{N} \log \frac{\exp(s_{ii})}{\sum_{j=1}^{N} \exp(s_{ij})}, 
    L_{a \to v} &= -\frac{1}{N} \sum_{i=1}^{N} \log \frac{\exp(s_{ii})}{\sum_{j=1}^{N} \exp(s_{ji})}.
\end{aligned}
\end{equation}

The final objective for the \alignmodelname\ module, $L_{\text{o-align}}$, is the average of these two directional losses, encouraging a bidirectional alignment between the modalities:
$L_{\text{o-align}} = \frac{1}{2}(L_{v \to a} + L_{a \to v})$.

While \alignmodelname effectively aligns the high-level semantics of visual and audio embeddings, it falls short in modeling their temporal relationships. To overcome this limitation, we introduce two techniques: Temporal Embedding Grouping and Constrained Rotary Time Embedding, which are described in the following sections.

\noindent\textbf{Temporal Embedding Grouping (TEG).}
We first impose temporal order to visual-audio embeddings by organizing them into groups based on their timestamps. The relative temporal order information is then encoded in the position of visual and audio embeddings in the input sequence.

Let the duration of each temporal group be $T_G$, which controls the granularity of the grouping. For simplicity, consider a case where we only sample {four} visual frames at timestamps $\{t^1_{v}, t^2_{v}, t^3_{v}, t^4_{v}\}$ and four audio samples at timestamps $\{t^1_{a}, t^2_{a}, t^3_{a}, t^4_{a}\}$. These timestamps satisfy $t_v^1 < t_v^2 < T_G < t_v^3 < t_v^4 < 2T_G$ and $t_a^1 < t_a^2 < T_G < t_a^3 < t_a^4 < 2T_G$. The corresponding set of visual embeddings is $E_v = \{\mathbf{e}_{v}^{t_v^1}, \mathbf{e}_{v}^{t_v^2}, \mathbf{e}_{v}^{t_v^3}, \mathbf{e}_{v}^{t_v^4}\}$, where each embedding $\mathbf{e}_{v} \in \mathbb{R}^{(HW)\times C}$. Here, $H$ and $W$ represent the height and width of the visual feature map, and $C$ is the latent dimension. Similarly, the set of audio embeddings is $E_a = \{\mathbf{e}_{a}^{t_a^1}, \mathbf{e}_{a}^{t_a^2}, \mathbf{e}_{a}^{t_a^3}, \mathbf{e}_{a}^{t_a^4}\}$, with each $\mathbf{e}_{a} \in \mathbb{R}^{1 \times C}$. Based on their timestamps relative to the duration $T_G$, the embeddings for each modality are partitioned into two temporal groups:
\begin{equation}
\begin{aligned}
G^1_v = \{\mathbf{e}_{v}^{t_v^1}, \mathbf{e}_{v}^{t_v^2}\},  G^2_v = \{\mathbf{e}_{v}^{t_v^3}, \mathbf{e}_{v}^{t_v^4}\}, 
G^1_a = \{\mathbf{e}_{a}^{t_a^1}, \mathbf{e}_{a}^{t_a^2}\},  G^2_a = \{\mathbf{e}_{a}^{t_a^3}, \mathbf{e}_{a}^{t_a^4}\}.
\end{aligned}
\end{equation}

Then we combine the visual and audio groups based on temporal order, and obtain the omni-modal embedding sequence:
\begin{equation}
\mathbf{E}_{\text{group}} = \left[G^1_v, G^1_a, G^2_v, G^2_a\right]= 
\left[ \mathbf{e}_{v}^{t_v^1}, \mathbf{e}_{v}^{t_v^2}, \mathbf{e}_{a}^{t_a^1}, \mathbf{e}_{a}^{t_a^2}, \mathbf{e}_{v}^{t_v^3}, \mathbf{e}_{v}^{t_v^4}, \mathbf{e}_{a}^{t_a^3}, \mathbf{e}_{a}^{t_a^4}\right].
\end{equation}

This temporal organization of the embedding sequence allows the subsequent LLM backbone to better capture the temporal relationships among embeddings from different modalities. Our experiments show that this time-based grouping improves the model's ability to comprehend information from multiple modalities.

\noindent\textbf{Constrained Rotary Time Embedding (CRTE).}
TEG incorporates relative temporal order into embeddings but still lacks the ability to encode absolute timestamp information. Prior work, RoTE~\citep{goel2024omcat}, explored embedding rotations to inject absolute timestamps, but this method remains sensitive to minor timestamp fluctuations and struggles to capture larger temporal shifts effectively. To overcome these limitations, we introduce a constrained timestamp embedding strategy that defines a maximum time horizon, $T_{\text{max}}$, enabling a more balanced temporal sensitivity. Our approach comprises three stages: base frequency construction, frequency modulation, and element-wise rotary embedding, as described next.

\noindent\textbf{Base Frequency Generation.}
We first define base frequencies as:
\begin{equation}
\omega_i = \frac{2\pi}{T_{\text{max}} \theta^{i/C}}, \quad \text{for} \quad i = 0, 1, \ldots, C-1,
\end{equation}
where $\omega_i$ is the base frequency for dimension $i$, $C$ is the embedding dimension, $\theta \geq 1$ controls frequency scaling, and $T_{\text{max}}$ defines the coarsest temporal resolution. A smaller $T_{\text{max}}$ increases frequency and sensitivity to fine-grained differences, while a larger one captures broader trends but may blur close timestamps, and is thus critical for balancing local and global temporal encoding. 

\noindent
\textbf{Frequency Modulation.}
To adapt frequencies to actual timestamps, we scale them as: $\Omega_{i,j} = \omega_i \cdot t_j$,
where $\Omega_{i,j}$ is the modulated frequency at dimension $i$ and time $t_j$ for sample $j$. This step ensures that temporal differences are reflected in the rotation applied to embeddings. 

\noindent
\textbf{Rotary Embedding Application.}
Similar to RoPE~\citep{su2024roformer}, given an embedding vector $\mathbf{x} \in \mathbb{R}^C$ of sample $j$ (a sampled frame for video or a sampling point for audio), we apply rotation as:
\begin{equation}
\text{CRTE}(\mathbf{x}, \Omega_{:,j}) = \mathbf{x} \odot \cos(\Omega_{:,j}) + \texttt{RotateHalf}(\mathbf{x}) \odot \sin(\Omega_{:,j}),
\end{equation}
where $\odot$ denotes element-wise multiplication, and $\texttt{RotateHalf}$ rotates each pair of dimensions by $90^0$:
$\texttt{RotateHalf}(\mathbf{x}) = [-x_2, x_1, -x_4, x_3, \ldots, -x_C, x_{C-1}].$
The $\texttt{RotateHalf}$ function effectively groups the entire $C$-dimensional embedding vector into $C/2$ independent 2D planes. Each of these 2D planes gets its own rotation, and the angle of rotation can be different for each pair.
We apply rotations at varying frequencies across different pairs of dimensions for two primary reasons: it enables a rich, multi-scale representation of temporal information, and it preserves the semantic integrity of the original embedding vectors.

The base frequency in CRTE, $\omega_i$ is designed to have a geometric progression of frequencies.
For small values of $i$ (\textit{e.g.}, the first pairs of dimensions), the denominator is smaller, resulting in higher frequencies ($\omega_i$ is large). These dimensions undergo rapid rotation with respect to time. Consequently, they are highly sensitive to fine-grained temporal differences and are effective at distinguishing between timestamps that are close to one another.
For large values of $i$ (\textit{e.g.}, the last pairs of dimensions), the term $\theta^{i/d}$ becomes significantly larger, resulting in lower frequencies ($\omega_i$ is small). These dimensions rotate slowly, making them suitable for encoding coarse, long-range temporal relationships. They provide a stable signal for large time intervals without the issue of aliasing or ``wrapping around'' that would occur with high-frequency signals.
By partitioning the embedding space into a spectrum of frequencies, the model can concurrently attend to both local and global temporal contexts. This multi-scale approach provides a robust and comprehensive representation of absolute time.

\noindent
\textbf{Final Embedding Sequence.}
After CRTE, the temporally-aligned omni-modal embedding sequence is passed into the LLM backbone, allowing it to integrate both fine- and coarse-grained timing cues during downstream processing.

\noindent\textbf{Input-Output Configuration.}
The final architecture perceives flexible input modality combinations with a subset or union of all modalities, \textit{e.g.}, video with or without audio, with speech or text prompts. On the output end, the text-output based system can be connected with off-the-shelf Text-to-Speech~(TTS) modules -- we analyze their tradeoffs in Section~\ref{sec:exp_speech_output}. Without bells and whistles, users can generate spoken descriptions for videos, answer spoken questions, or verbally instruct robots.

\section{Training Strategy}

To gradually enable comprehensive omni-modal understanding of a pretrained LLM, we use a two-stage approach: we first conduct modality-specific training to develop individual capabilities for each modality, followed by omni-modal joint training to integrate these capabilities.

\subsection{Modality-Specific Training}
Starting from a pretrained LLM, Qwen2.5-7B-Instruct model~\citep{qwen25}, we begin by developing the model's capabilities for vision and audio comprehension independently, utilizing data tailored to each modality (\textit{i.e.}, those containing only visual or only auditory information). 
Due to space limitations, we present comprehensive details of this phase in Appendix~\ref{sec:modality_specific_train} and proceed directly to describe the subsequent omni-modal joint training phase in the next section.

\subsection{Omni-Modal Joint Training}

\label{sec:data}

We employ two types of data in the omni-modal joint training phase: (i) modality-specific data, randomly sampled from the datasets used in the earlier vision-only and audio-only training, and (ii) omni-modal data, which contains both vision and audio inputs.
For the omni-modal data, which contains both visual and audio inputs, can be further divided into two categories, \textit{i.e.}, implicit omni-modal learning data and explicit omni-modal learning data, depending on how the omni-modal understanding ability is supervised in training.

\noindent\textbf{(i) Implicit Learning Data.}
Videos are naturally omni-modal when visual and audio streams are present simultaneously but remains under explored. We first take advantage of the existing video QA datasets to supervise the visual-audio joint understanding ability implicitly, which is underutilized in most previous video LLMs. This practice, we refer as \textit{implicit omni-modal learning}, leads to notably improved performance in video understanding that remains under utilized by prior work.

\noindent\textbf{(ii) Explicit Learning Data.}
To obtain more direct and accurate supervision for joint visual-audio understanding ability, we further propose an omni-modal data engine to synthesize omni-modal labeling for videos with audio tracks, enabling us to conduct \textit{explicit omni-modal learning}.

\begin{figure*}[t]
\centering
% \vspace{-30pt}
\includegraphics[width=1.\linewidth]{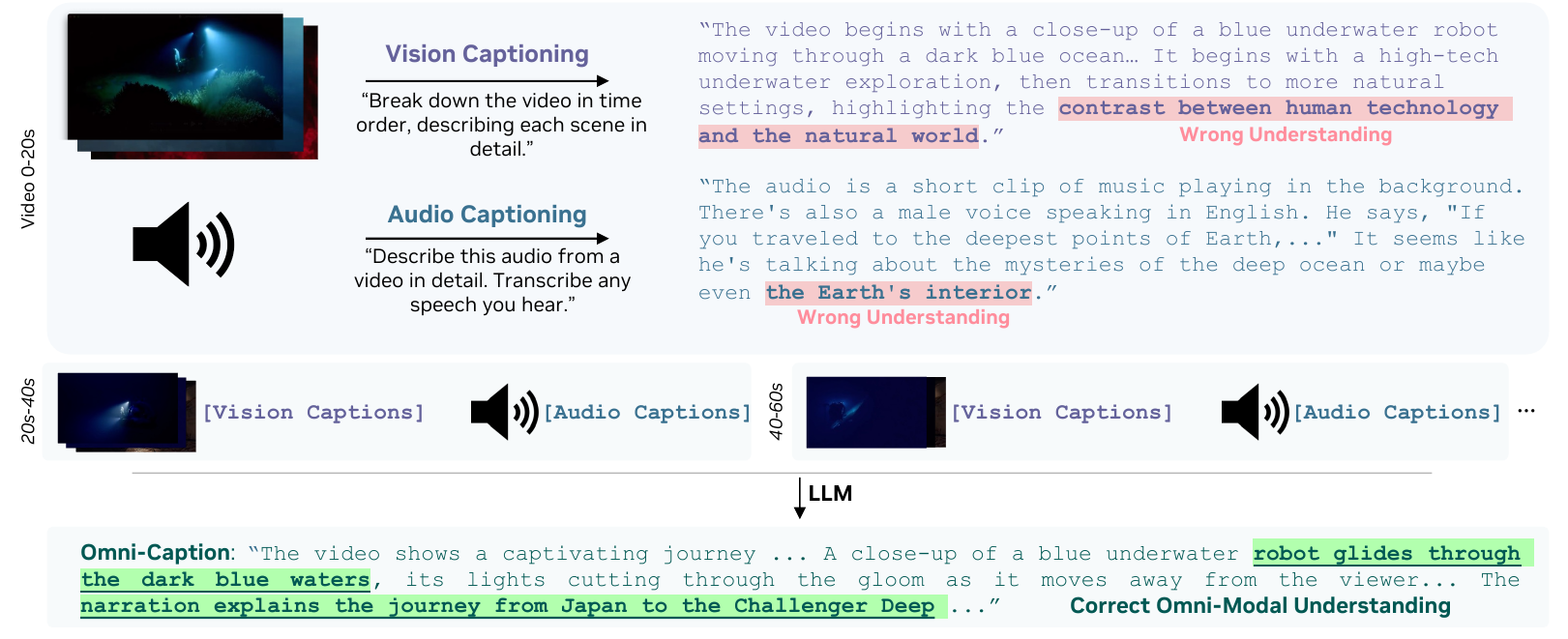}
% \vspace{-20pt}
\caption{Omni-modal captions generation pipeline.
Video is segmented into 20-second clips. Visual and audio captions are generated independently for each segment, but lack cross-modal context and contain wrong understanding (modality-specific hallucination). A separate LLM performs cross-modal correction and summarization to create accurate omni-modal captions. 
}
% \vspace{-10pt}
\label{fig:omni_caption}
\end{figure*}

\noindent\textbf{Omni-Modal Data Engine.} The whole data engine is visualized in Figure~\ref{fig:omni_caption}. We start with synthetic audio and video captions using pretrained vision captioning model~\citep{zhu2025internvl3} and audio captioning model~\citep{xu2025qwen25omni}.
We immediately observed that captions generated from either modality alone can lead to wrong understanding due to the inherent modality-specific limitations. As illustrated in Figure~\ref{fig:omni_caption}, the video is centered around deep-sea exploration. However, the vision-captioning model incorrectly interpreted it as being only about human technology, relying solely on visual cues without access to the speech in video. Conversely, the audio-captioning model wrongly labeled it as related to ``Earth’s interior”, since it could only draw meaning from the audio track. We refer to this limitation as ``\textit{modality-specific hallucination}''. To address this issue, we employ a LLM~\citep{yang2025qwen3} to correct and summarize the visual and audio captions based on information from both sides, producing a comprehensive joint caption for each 2-minute segment. From our observation, this method can help achieve correct omni-modal understanding, as shown in the example in Figure~\ref{fig:omni_caption}.
Furthermore, we enhance the diversity and quality of the omni-modal data by synthesizing QA pairs with reasoning trace from the omni-modal captions using a reasoning LLM~\citep{guo2025deepseek}. The resulting dataset greatly assists with learning, as we show in experiments.

\begin{tcolorbox}[highlightstyle]
\keyinsight{}
Captioning based solely on audio or visual is often inaccurate because of the inherent limitations of each modality. Hence, a joint captioning approach is preferred to integrate both modalities and produce comprehensive summaries across clips.
\end{tcolorbox}

\begin{figure*}[t]
\centering
% \vspace{-40pt}
\includegraphics[width=\linewidth]{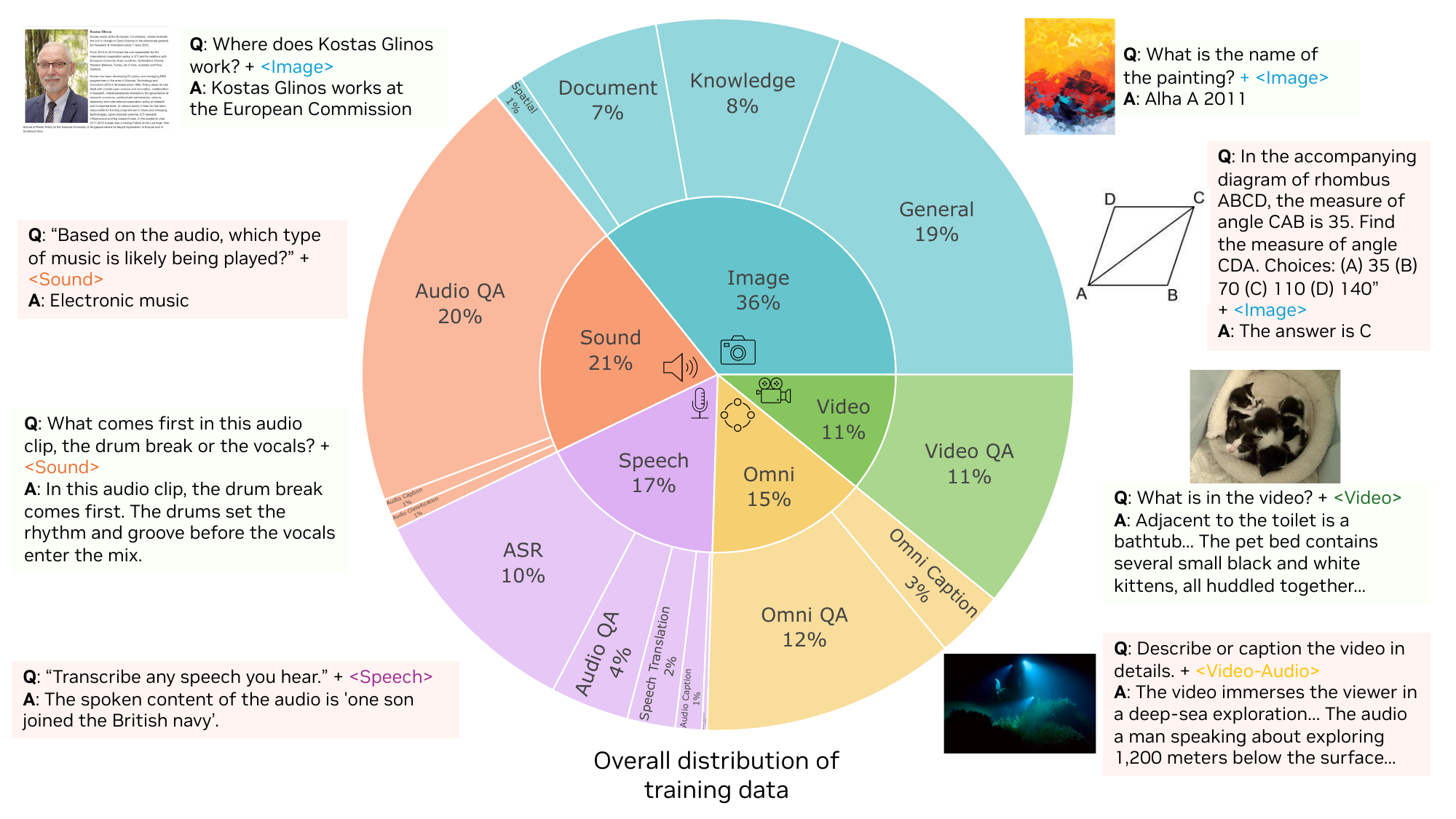}
\vspace{-20pt}
\caption{Pie chart of overall distribution of training data across modalities, showing proportions for image (36\%), non-speech sound (21\%), speech (17\%), omni (15\%), and video (11\%). 
}
% \vspace{-10pt}
\label{fig:data_dist}
\end{figure*}

\noindent\textbf{Joint Training Data Distribution.}
As shown in Figure~\ref{fig:data_dist}, the overall training dataset contains 24 million modality-specific conversations from 150+ sub-datasets across image, video, and audio understanding tasks.
Omni-modal data contributes 15\%, image data constitutes the largest share at 36\%, speech data represents 17\% of the total, and video data forms the remaining 11\%. 
For more details, please refer to Appendix~\ref{sec:detail_training_data}. To enable audio-prompted ability, we convert text prompts in multimodal tasks into speech using Magpie TTS model \citep{hussain2025koel,neekhara2024improving,casanova2025low}, generating omni-modal speech-visual input pairs. The questions are generated from a comprehensive collection of multimodal datasets, including general multimodal understanding, image captioning, spatial relationship reasoning and referring, chart and table interpretation, scientific figure analysis, document understanding, and multi-hop reasoning. This diverse range enables comprehensive evaluation across core vision-language capabilities such as factual grounding, reasoning over structured data, and complex multi-step inference in both scientific and general domains. See detailed distribution of speech-prompted omni QA datasets in Figure~\ref{fig:speech_vqa_data}.

\section{Experiments}
\label{sec:experiments}
We start with ablations to validate our design options in Section~\ref{sec:exp_ablation}, before large-scale training towards frontier performances in Section~\ref{sec:exp:large_scale}.

\begin{table}[ht]
  \centering
  % \vspace{-\baselineskip}
  \caption{Ablation study for omni-modal alignment. The proposed Temporal Embedding Grouping (TEG), Constrained Rotary Time Embedding (CRTE), and \alignmodelname consistently achieve better average performance across modalities.}
  \vspace{-10pt}
  \resizebox{.7\linewidth}{!}{
  \huge
  \tablestyle{2.5pt}{1.2}
  \begin{tabular}{llllll}
    \toprule
    \multicolumn{1}{c}{\multirow{2}{*}{\textbf{Method}}} & \multicolumn{4}{c}{\textbf{Omni}} \\
    & Worldsense $\uparrow$ & Dailyomni $\uparrow$ & Omnibench $\uparrow$ & Average $\uparrow$ \\
    \midrule
    Token Concatenation -- Baseline & 42.21 & 54.55 & 36.46 & 45.51 \\
    \midrule
    + TEG \textbf{(ours)} & $44.51_{\textcolor{deepgreen}{+2.30}}$ & $60.99_{\textcolor{deepgreen}{+6.44}}$ & $37.65_{\textcolor{deepgreen}{+1.19}}$ & $47.72_{\textcolor{deepgreen}{+2.21}}$ \\
    \midrule
    ++ Learned Time Embedding & $44.58_{\textcolor{deepgreen}{+2.37}}$ & $60.40_{\textcolor{deepgreen}{+5.85}}$ & $36.91_{\textcolor{deepgreen}{+0.45}}$ & $47.30_{\textcolor{deepgreen}{+1.79}}$ \\
    ++ RoTE & $44.42_{\textcolor{deepgreen}{+2.21}}$ & $60.74_{\textcolor{deepgreen}{+6.19}}$ & $38.24_{\textcolor{deepgreen}{+1.78}}$ & $47.80_{\textcolor{deepgreen}{+2.29}}$ \\
    ++ CRTE (\textbf{ours}) & $45.46_{\textcolor{deepgreen}{+3.25}}$ & $65.66_{\textcolor{deepgreen}{+11.11}}$ & $39.64_{\textcolor{deepgreen}{+3.18}}$ & $50.25_{\textcolor{deepgreen}{+4.74}}$ \\
    \midrule
    \rowcolor{lightgreen}
    +++ \alignmodelname (\textbf{ours}) & $\textbf{46.21}_{\textcolor{deepgreen}{+4.00}}$ & $\textbf{65.83}_{\textcolor{deepgreen}{+12.28}}$ & $\textbf{45.74}_{\textcolor{deepgreen}{+9.28}}$ & $\textbf{52.59}_{\textcolor{deepgreen}{+7.08}}$ \\
    \bottomrule
  \end{tabular}
  }
\label{tab:embed_rotate}
\end{table}

\subsection{Design Choice Ablation}
\label{sec:exp_ablation}

\subsubsection{Visual-Audio Alignment Scheme}
\textbf{Baseline Setup.}
To investigate the behavior of omni-modal models under various experimental conditions we gradually introduce new techniques onto a baseline model trained with 10B tokens randomly sampled subset of the full data mixture (the sampling process is weighted according to the original dataset sizes). We evaluate model performance on Worldsense~\citep{benchekroun2023worldsense}, Dailyomni~\citep{zhou2025daily}, and Omnibench~\citep{li2024omnibench}.

\noindent\textbf{Temporal Embedding Grouping.}
We observe immediate performance improvements with TEG applied to the baseline and present the results in Table~\ref{tab:embed_rotate}, thanks to the enhanced temporal alignment of modality tokens.

\noindent\textbf{Constrained Rotary Time Embedding.}
We next compare CRTE with other design choices: 
\textit{(i) ``Learned Time Embedding''} that defines a trainable embedding matrix, where each discrete timestamp in the range $[0, T_{max}]$ is mapped to a unique vector via MLP. 
\textit{(ii) ``RoTE''~\citep{goel2024omcat}}, a recent embedding method introduced in Section~\ref{sec:omni_align_mechanism}. 
As summarized in Table~\ref{tab:embed_rotate}, the ``Learned Time Embedding'' method slightly degrades performance (47.30), indicating it is unsuitable for absolute timestamps. RoTE offers only marginal gains, while the proposed Constrained Rotary Time Embedding achieves the best score (50.25), clearly improving over the baseline.

\noindent\textbf{\alignmodelname.}
Finally, we impose the proposed OmniAlignNet on top of TEG and CRTE. As shown in the bottom section of Table~\ref{tab:embed_rotate}, OmniAlignNet delivers significant performance boosts across all benchmarks. The average score improves from 50.25 to 52.59 (+2.34), and the model achieves considerable gains on {Omnibench} (+6.1), {Worldsense} (+0.75), and {Dailyomni} (+1.17).

\subsubsection{Implicit and Explicit Learning}
\label{sec:implicit_explicit_exp}

\begin{table}
  \centering
  % \vspace{-\baselineskip}
  \caption{Ablation study on joint visual-audio learning methods. ``Visual+Audio” uses audio in video for implicit learning (IL), while  ``data engine'' generates omni-modal data for explicit learning (EL). }
  \vspace{-10pt}
  \tablestyle{4pt}{1.}
  \resizebox{.9\linewidth}{!}{
  % \huge
  \begin{tabular}{llllllccc}
    \toprule
    \multicolumn{1}{c}{\multirow{2}{*}{\textbf{Method}}} &
    \multicolumn{2}{c}{\textbf{VideoMME $\uparrow$}} &
    \multicolumn{3}{c}{\textbf{VideoMME w/o sub. $\uparrow$}} \\
    \cmidrule(lr){2-3} \cmidrule(lr){4-6}
    & w/ subtitles & w/o subtitles & Short & Medium & Long \\
    \midrule
    Visual Alone & 66.37 & 61.67 & 74.22 & 59.67 & 51.11 \\
    Visual + Audio (IL)  
      & 66.96\textsubscript{\textcolor{deepgreen}{+0.59}}
      & 63.76\textsubscript{\textcolor{deepgreen}{+2.09}}
      & 71.31\textsubscript{\textcolor{deepgreen}{-2.91}}
      & 64.16\textsubscript{\textcolor{deepgreen}{+4.49}}
      & 55.82\textsubscript{\textcolor{deepgreen}{+4.71}} \\
    \rowcolor{lightgreen}
    Visual + Audio + Data Engine (EL)  
      & \textbf{68.63}\textsubscript{\textcolor{deepgreen}{+2.26}}
      & \textbf{67.37}\textsubscript{\textcolor{deepgreen}{+5.70}}
      & \textbf{76.78}\textsubscript{\textcolor{deepgreen}{+2.56}}
      & \textbf{67.56}\textsubscript{\textcolor{deepgreen}{+7.89}}
      & \textbf{57.78}\textsubscript{\textcolor{deepgreen}{+6.67}} \\
    \bottomrule
  \end{tabular}
  }
  \label{tab:implicit_explicit}
\end{table}

We next validate implicit and explicit omni-modal learning as detailed in Section~\ref{sec:data}. For implicit learning, we continue to finetune the above model on ~270K video conversations with audio stream. Results in Table~\ref{tab:implicit_explicit} show clear gains on VideoMME~\citep{fu2024videomme}, even when subtitles are provided, highlighting the value of learning directly from audio. Further adding explicit learning data from our omni-modal data engine yields stronger improvements across benchmarks, showing the effectiveness of our data pipeline.

\subsection{Scaling and Evaluation}
\label{sec:exp:large_scale}
With validated design choices, we now scale up the experiments using the full post-training omni-modal dataset introduced in Section~\ref{sec:data}. Training details are in Appendix~\ref{sec:detail_joint_train}.

\begin{table}[h]
  \centering
  % \vspace{-\baselineskip}
  \caption{Omni benchmarks, including video–audio datasets Worldsense and Dailyomni, as well as the image–audio dataset Omnibench.}
  \vspace{-10pt}
  \resizebox{.65\linewidth}{!}{
  \huge
  \tablestyle{2pt}{1.2}
  \begin{tabular}{lcccc}
    \toprule
    \multirow{2}{*}{\textbf{Model}} 
      & \multicolumn{4}{c}{\textbf{Omni}} \\
    & \makecell{Worldsense \\ (\textit{Video–Audio} $\uparrow$)} 
    & \makecell{Dailyomni \\ (\textit{Video–Audio} $\uparrow$)} 
    & \makecell{Omnibench \\ (\textit{Image–Audio} $\uparrow$)} 
    & \makecell{Avg. \\ ($\uparrow$)} \\
    \midrule
    Gemini           & --         & 61.32 (2.0 Flash Lite) & 42.91 (1.5 Pro) & - \\
    GPT-4o           & 42.60       & --        & -- & - \\
    \hline
    InternVL2        & 39.10       & --                     & 47.55 (v2.5) & - \\
    Qwen2-VL         & 32.40       & --                     & 48.60 & - \\
    Qwen2.5-Omni     & \underline{45.40} & 47.45 & \textbf{56.13} & 49.66 \\
    \midrule
    \rowcolor{lightgreen}
    \textbf{\modelname}    & \textbf{48.23}      & \textbf{66.50}         & 46.47 & \textbf{53.73} \\
    \bottomrule
  \end{tabular}
  }
  \label{tab:omni_results}
\end{table}

\subsubsection{Omni-Modal Benchmark}
We first evaluate on omni-modal understanding benchmarks and show results in Table~\ref{tab:omni_results}.
\modelname sets a new state-of-the-art average score of {53.73}, and marks a notable improvement of \textbf{+4.07} compared to the next best model, Qwen2.5-Omni.
On the {Worldsense} benchmark, our model achieves the highest score of {48.23}, surpassing Qwen2.5-Omni by \textbf{+2.83}. The advantage is even more significant on the {Dailyomni} dataset, where our model attains a score of {66.50}, leading by \textbf{+19.05} over Qwen2.5-Omni and by \textbf{+5.18} over Gemini-2.0-Flash-Lite.
In the {Omnibench} benchmark, our model shows a solid score of {46.47}, higher than Gemini 1.5 Pro.
\begin{wraptable}{r}{0.32\linewidth}
\centering
% \huge
\vspace{19pt}
\caption{Audio QA benchmark.
}
\vspace{-10pt}
\resizebox{\linewidth}{!}{%
\tablestyle{4pt}{1.2}
\begin{tabular}{cc}
\toprule
\textbf{Model} & \textbf{MMAR} ($\uparrow$) \\
\hline
LTU & 19.20 \\
Audio Flamingo 2 & 21.90 \\
Qwen-2-Audio & 30.40 \\
SALAMONN & 33.20 \\
Baichuan-Omni-1.5 & 40.70 \\
Qwen2.5-Omni   & 56.70 \\
\midrule
 \rowcolor{lightgreen}
\textbf{\modelname}   & \textbf{58.40} \\
\bottomrule
\label{tab:mmar_scores}
\end{tabular}
}
\vspace{-50pt}
\end{wraptable}

\subsubsection{Audio Benchmark}
\noindent\textbf{Audio QA.}
We assess our model 
on audio understanding benchmarks, 
MMAR~\citep{ma2025mmar} and MMAU~\citep{sakshi2024mmau}, 
with results reported in Tables~\ref{tab:mmar_scores} and \ref{tab:mmau_scores}. 
On MMAR, \modelname surpasses Qwen2.5-Omni by +1.7, and on MMAU by +0.6, highlighting significant improvement in general audio understanding.

\noindent\textbf{Speech Recognition.}
To assess the automatic speech recognition~(ASR) capabilities of \modelname, we evaluate it on four widely used benchmarks: LibriSpeech~\citep{panayotov2015librispeech}, AMI~\citep{kraaij2005ami}, Tedlium~\citep{rousseau2012ted}, and VoxPopuli\citep{wang2021voxpopuli}, comparing against leading multi-modal models. As shown in Table~\ref{tab:asr_scores}, our model achieves competitive word error rates (WER) of {1.7} on LibriSpeech-clean and {3.7} on LibriSpeech-other, closely matching or surpassing the latest works.

\begin{table}
  \centering
    % \vspace{-\baselineskip}
  \caption{Multi-domain speech recognition benchmarks. $^*$Results taken from related papers; details in Appendix~\ref{appendix:e2:asr}.}
  \vspace{-10pt}
  \tablestyle{6pt}{1.2}
  \resizebox{.63\linewidth}{!}{%
  \vspace{-10pt}
  \begin{tabular}{lccccc|cc}
    \toprule
    \multicolumn{1}{c}{\multirow{2}{*}{\textbf{Model}}} 
      & \multicolumn{6}{c}{\textbf{WER ($\downarrow$)}} \\
    & LS$_{\text{clean}}$ & LS$_{\text{other}}$ 
      & {AMI} & {Ted.} & {Vox.} & {Avg.}  \\
    \midrule
    Whisper-large-v3      & 1.8 & {3.6} & 16.1 & 3.9 & 10.1 & 7.1  \\
    Qwen2-Audio          & 1.7 & 4.1 & 15.2 & 3.1 & 7.1 & 6.4  \\
    GPT-4o-real-time     & 2.5 & 5.0 & 19.3 & 4.1 & 12.1 & 8.6  \\
    Gemini-2.0-Flash     & 2.5 & 5.9 & 21.5 & 3.0 & 7.9 & 8.2  \\
    Phi-4-MM             & \textbf{1.7} & 3.8 & \textbf{11.5} & \textbf{2.9} & 5.9 & \textbf{5.2}  \\
    Qwen2.5-omni         & 1.8$^*$ & \textbf{3.4}$^*$ & 17.9 & 5.2 & \textbf{5.8}$^*$ & 6.8  \\
    \midrule
    \rowcolor{lightgreen}
    \textbf{\modelname}         & \textbf{1.7} & 3.7 & 16.1 & 3.4 & 6.8 &  6.3 \\
    \bottomrule
  \end{tabular}
}
  \label{tab:asr_scores}
\end{table}

We further investigate \modelname's performance under two agentic-cascaded setups: (i) incorporating ASR text history~\citep{huang2025step} and (ii) leveraging retriever-based training as shown in Figure~\ref{fig:asr:test-time}. These techniques help boost \modelname's capacity, yielding average WERs of \textbf{5.7} and \textbf{5.0}, respectively. These test-time scaling studies are provided in Appendix~\ref{sec:rag} (Table~\ref{tab:asr_appendix}).

\subsubsection{Video Benchmark}

\begin{table}[t]
  \centering
  \caption{MMAU audio benchmark.}
    \vspace{-10pt}
  \tablestyle{6pt}{1.2}
  \resizebox{.95\linewidth}{!}{%
  \begin{tabular}{lcccccc|cc}
        \toprule
        \multicolumn{1}{c}{\multirow{2}{*}{\textbf{Model}}} &
        \multicolumn{2}{c}{\textbf{Music}} &
        \multicolumn{2}{c}{\textbf{Sound}} &
        \multicolumn{2}{c}{\textbf{Speech}} &
        \multicolumn{2}{c}{\textbf{Avg}} \\
        & \textbf{Test} & \textbf{Test-mini} &
        \textbf{Test} & \textbf{Test-mini} &
        \textbf{Test} & \textbf{Test-mini} &
        \textbf{Test} & \textbf{Test-mini} \\
        \midrule
        Gemini 2.5 Pro & 68.26 & 64.77 & 70.63 & 75.08 & 72.67 & 71.47 & 71.60 & 69.36 \\
        Gemini 2.5 Flash & 76.58 & 69.40 & 65.57 & 69.50 & 71.80 & 68.27 & 69.57 & 67.39 \\
        Kimi-Audio & 62.16 & 65.93 & 66.77 & 70.70 & 56.57 & 68.20 & 64.40 & 68.20 \\
        Phi-4-multimodal & 61.97 & 64.37 & 62.67 & 65.47 & 63.80 & 67.27 & 62.81 & 65.70 \\
        Audio Flamingo 2 & 44.74 & 70.20 & 68.13 & 70.96 & 44.87 & 62.40 & 61.06 & 62.40 \\
        GPT-4o Audio & 49.93 & 56.29 & 63.20 & 64.56 & 69.33 & 66.67 & 60.82 & 62.50 \\
        Qwen2-Audio-Instruct & 55.26 & 55.67 & 56.29 & 61.17 & 59.60 & 55.37 & 57.40 & 59.60 \\
        Gemma 3n 4B & 61.26 & 53.20 & 56.89 & 50.27 & 58.00 & 62.13 & 58.00 & 55.20 \\
        Qwen2.5-Omni & 67.33 & 65.90 & 76.77 & 78.10 & 68.90 & 70.60 & 71.00 & 71.50 \\
    \rowcolor{lightgreen}
    \textbf{\modelname} & 73.07 & 73.65 & 73.57 & 78.68 & 68.17 & 66.97 & \textbf{71.60} & \textbf{73.10} \\
    \bottomrule
        \end{tabular}}
        \label{tab:mmau_scores}
\end{table}

We compare with other open-source video-language models in Table~\ref{tab:results:video}. 
On the LongVideoBench~\citep{wu2024longvideobench} val set, {\modelname} achieves a score of 61.3, outperforming NVILA by a margin of \textbf{+3.6}. Similarly, our model improves on MVBench~\citep{li2024mvbench} with a score of 70.6, outperforming also the recently released Qwen2.5-Omni (70.3). 

Furthermore, on the Video-MME~\citep{fu2024videomme} benchmark (without subtitles hints), {\modelname} again sets a high score at 68.2, surpassing Qwen2.5-VL-7B by \textbf{+3.1}, positioning it as a leading open-source model for video comprehension tasks.

\begin{tcolorbox}[highlightstyle]
\keyinsight{}
Audio understanding capacity enables consistent metric improvements across video benchmarks, akin to human perception. 
\end{tcolorbox}

\subsubsection{Image Benchmark}
\begin{table*}[t]
    \centering
        \centering
        \caption{Video benchmarks. \modelname outperforms NVILA baseline.
        }
        \vspace{-10pt}
       \tablestyle{4pt}{1.2}
        \resizebox{.68\linewidth}{!}{%
        \begin{tabular}{lrcccc}
            \toprule
             \multicolumn{2}{c}{\multirow{2}{*}{\textbf{Model}}}  & \multicolumn{2}{c}{\textbf{LongVideoBench $\uparrow$}} & \multicolumn{1}{c}{\textbf{MVBench $\uparrow$}} & \multicolumn{1}{c}{\textbf{Video-MME $\uparrow$}} \\
            \cmidrule(lr){3-4}\cmidrule(lr){5-5}\cmidrule(lr){6-6} 
            & & val & test & test & w/o sub. \\
            \midrule
            GPT-4o mini & - & 56.5 & 58.8 & -- & 64.8 \\
            GPT-4o & - & 66.7 & 66.7 & --  & 71.9\\
            \midrule
            LLaVA-NeXT-Video & 7B & 43.5 & 43.5 & 33.7  & 46.5 \\
            InternVL2 & 8B & 54.6 & -- & 65.8 & 56.3 \\
            LLaVA-OneVision & 8B & 56.5 & -- & 56.7 & 58.2 \\
            LongVILA & 7B & 57.1 & -- & 67.1 & 60.1 \\
            Qwen2.5-VL &  8B & 56.0  & - & {69.6} & {65.1}  \\
            {Qwen2.5-Omni} &  11B &  - & - & \underline{70.3} & 64.3  \\
            \midrule
            NVILA & 8B & \underline{57.7} & \underline{58.7} & {68.1} & {64.2} \\
            \midrule    
            \rowcolor{lightgreen}
            \textbf{\modelname} & 9B & \textbf{61.3}  & \textbf{62.0}   & \textbf{70.6}  & \textbf{68.2} \\
            \bottomrule
        \end{tabular}
        }
        \label{tab:results:video}
    \end{table*}
    
    \begin{table}[t]
        \centering
        \caption{Image benchmarks. \modelname maintains comparable image understanding performance with NVILA.}
        \vspace{-10pt}
        \tablestyle{2pt}{1.2}
        \resizebox{\linewidth}{!}{%
        \begin{tabular}{lrcccccccccccc}
            \toprule
            & & AI2D & ChartQA & DocVQA & InfoVQA & MathVista & \multicolumn{3}{c}{MMMU} & \multirow{2.5}{*}{\makecell{Real- \\ WorldQA}} & SEED & TextVQA & VQAv2 \\
            \cmidrule(lr){3-3}\cmidrule(lr){4-4}\cmidrule(lr){5-5}\cmidrule(lr){6-6}\cmidrule(lr){7-7}\cmidrule(lr){8-10}\cmidrule(lr){12-12}\cmidrule(lr){13-13}\cmidrule(lr){14-14}
            & & test & test & test & test & testmini & val & test & pro & & image & val & testdev \\
            \midrule
            GPT-4o & -- & 94.2 & 85.7 & 92.8 & 79.2 & 63.8 & 69.1 & 64.7 & 51.9 & 75.4 & 76.2 & 77.4 & 78.7 \\
            Claude 3.5 Sonnet & -- & 94.7 & 90.8 & 85.2 & 74.3 & 67.7 & 68.3 & 63.7 & 51.5 & 60.1 & -- & 74.1 & 70.7 \\
            Gemini 1.5 Pro & -- & 94.4 & 87.2 & 93.1 & 81.0 & 63.9 & 62.2 & 57.6 & 43.5 & 70.4 & -- & 78.7 & 80.2\\
            \midrule
            LLaVA-1.5 & 7B & 55.5 & 17.8 & 28.1 & 25.8 & 25.6 & 35.7 & -- & -- & 54.8 & 66.1 & 58.2 & 78.5 \\
            VILA-1.5 & 8B & 76.6 & 52.7 & 40.6 & 25.9 & 36.7 & 38.6 & 32.7 & -- & 52.7 & 73.8 & 68.5 & 83.0	\\
            Cambrian-1 & 8B & 73.0 & 73.3 & 77.8 & 41.6 & 49.0 & 42.7 & -- & -- & 64.2 & 74.7 & 71.7 & 81.2 \\ 
            Florence-VL & 8B & 74.2 & 74.7 & 84.9 & 51.7 & 55.5 & 43.7 & -- & -- & 64.2 & 74.9 & 74.2 & 84.7 \\ 
            LLaVA-OneVision & 8B & 81.4 & 80.0 & 87.5 & 68.8 & 63.2 & 48.8 & 42.8 & 24.1 & 66.3 & 75.4 & 78.3 & 84.0 \\ 
            Llama 3.2 & 11B & \underline{91.9} & 83.4 & 88.4 & -- & 51.5 & 50.7 & -- & -- & -- & -- & -- & 75.2 \\ 
            InternVL2 & 8B & 83.8 & 83.3 & 91.6 & \underline{74.8} & 58.3 & \underline{51.2} & 42.6 & \underline{29.0} & 64.2 & 76.2 & 77.4 & 76.7 \\ 
            Qwen2-VL & 8B & 83.0 & 83.0 & \textbf{94.5} & \textbf{76.5} & 58.2 & \textbf{54.1} & \textbf{46.6} & \textbf{30.5} & \textbf{70.1} & 76.0 & \underline{84.3} & 82.9 \\ 
            \midrule
            NVILA & 8B & \textbf{92.3} & \textbf{86.1} & \underline{93.7} & 70.7 & 65.4 & 49.9 & 44.4 & 27.8 & \underline{68.6} & 76.5 & 80.1 & \textbf{85.4} \\ 
            \midrule
                \rowcolor{lightgreen}
            \textbf{\modelname} & 9B & 91.5 & 84.6 & 91.5 & 69.0 & 63.5 & 49.7 & 44.6 & 26.4 & 67.5 & \textbf{77.1} & 83.9 & \textbf{85.4} \\ 
            \bottomrule
        \end{tabular}
        }
        \label{tab:results:image}
    % \vspace{-10pt}
\end{table}

We evaluate {\modelname} on ten image benchmarks to test its versatility. 
These tasks range from understanding diagrams and charts (AI2D~\citep{kembhavi2016diagram}, ChartQA~\citep{masry2022chartqa}) to document analysis (DocVQA~\citep{mathew2021docvqa}), mathematics (MathVista~\citep{lu2024mathvista}) and general visual question answering (VQAv2-testdev~\citep{balanced_vqa_v2}). 
As shown in Table~\ref{tab:results:image}, \modelname consistently achieves competitive scores across the board.

\subsection{Omni-Modal Reasoning}
\label{sec:omni_reason}

Building on advances in the Group Relative Policy Optimization (GRPO)~\citep{shao2024deepseekmath} algorithm and prior work on multi-modal reasoning training~\citep{chen2025scaling,feng2025video}, we next tackle omni-modal reasoning through accommodating audio tokens in addition to visual ones.

Specifically, for each given question and omni-modal input $q=\{q_t, q_v, q_a\}$ ($q_t$ is textual input, $q_v$ is visual input, and $q_a$ is audio input, respectively),  the sampling number is $G$, the policy model, under the old policy $\pi_{\theta_{old}}$, generates a set of candidate answers $\{o_1, o_2,...,o_G\}$ along with corresponding rewards $\{r_1, r_2,...,r_G\}$, where the rewards are computed by a rule-based function that evaluates format and accuracy~\citep{shao2024deepseekmath}. The model $\pi_{\theta}$ is then optimized by maximizing the following objective:
\begin{align}\label{grpo}
\mathcal{J}(\theta) = \mathbb{E}_{q,\{o_i\}} [ \frac{1}{G} \sum_{i=1}^{G} (\mathrm{min}(\frac{\pi_{\theta}(o_i|q_t, q_v, q_a)}{\pi_{\theta_{old}}(o_i|q_t, q_v, q_a)}A_i, \mathrm{clip}(\frac{\pi_{\theta}(o_i|q_t, q_v, q_a)}{\pi_{\theta_{old}}(o_i|q_t, q_v, q_a)}, 1-\epsilon, 1+\epsilon)A_i) \nonumber\\- \beta\mathbb{D}_{KL}(\pi_\theta||\pi_{ref}))],
\end{align}

where $\epsilon$ and $\beta$ are hyper-parameters of each loss part, the sampling number $G$ is set as 8. The rewards $\{r_1, r_2,...,r_G\}$ are normalized to get the advantages ($A_i$) for updating the model:
\begin{equation}\label{eq2:grpo}
A_i = \frac{r_i - \mathrm{mean}(\{r_1, r_2,...,r_G\})}{\mathrm{std}(\{r_1, r_2,...,r_G\})}.
\end{equation}

\begin{table}
  \centering
  % \vspace{-\baselineskip}
  \caption{Ablation study of GRPO post-training.}
  \vspace{-10pt}
  \tablestyle{2.5pt}{1.2}
    \resizebox{.62\linewidth}{!}{%
  \begin{tabular}{lllll}
    \toprule
    \multicolumn{1}{c}{\multirow{2}{*}{\textbf{Model}}} & \multicolumn{4}{c}{\textbf{Omni}} \\
    & Worldsense ($\uparrow$) & Dailyomni ($\uparrow$) & Omnibench ($\uparrow$) & Avg. ($\uparrow$)\\
    \midrule
    \modelname    & 48.23 & 66.50 & 46.47 & 53.73 \\
    \rowcolor{lightgreen}
    \modelname \ + RL 
      & \textbf{48.70}\textsubscript{\textcolor{deepgreen}{+0.47}}
      & \textbf{67.08}\textsubscript{\textcolor{deepgreen}{+0.58}}
      & \textbf{47.79}\textsubscript{\textcolor{deepgreen}{+1.32}}
      & \textbf{54.52}\textsubscript{\textcolor{deepgreen}{+0.79}} \\
    \bottomrule
  \end{tabular}
  }
  \label{tab:grpo}
\end{table}

\begin{figure*}[t]
\centering
\includegraphics[width=\linewidth]{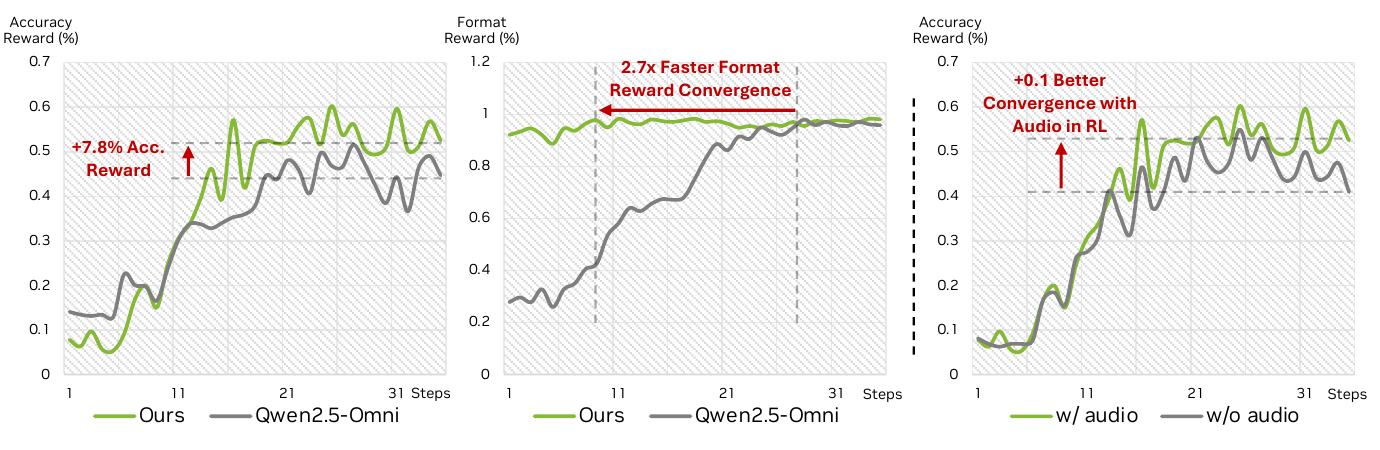}
\vspace{-20pt}
\caption{\textbf{Left:} Accuracy reward and format reward curves of \modelname and Qwen2.5-Omni in RL training. \textbf{Right:} Accuracy reward curve of \modelname with and without audio.
}
\vspace{-5pt}
\label{fig:rl_curve1}
\end{figure*}

We apply GRPO post-training to the final \modelname checkpoint to enhance its performance on omni-modal understanding benchmarks. For training data, we curated a 18K omni-modal MCQ dataset using the omni-modal data engine, as detailed in the methods section. During GRPO training, we utilize the Long-RL~\citep{chen2025scaling} as the training framework, configure the model to process up to 64 video frames, with a maximum prompt length of 1024 tokens and a maximum response length of 2048 tokens. The update batch size is set to 64, with the rollout number of 8 for each sample, ensuring robust gradient estimation. 
We employ a temperature of 1.0 and a top-p value of 0.99 for sampling, facilitating diverse exploration during training. 
These training configurations are carefully designed to optimize the model’s ability to handle complex omni-modal reasoning tasks effectively and efficiently.

As shown in Table~\ref{tab:grpo}, 
we observe consistent performance gains across all benchmarks after applying RL training. 
Comparing convergence with Qwen2.5-Omni under the same recipe (Figure~\ref{fig:rl_curve1}), both models benefit from our multi-modal RL framework, but \modelname leverages stronger base performance and instruction-following to surpass Qwen2.5-Omni on the GRPO accuracy curve within 15 steps, while also converging faster on formatting tasks. 
Ablation experiments further show that including audio input boosts RL effectiveness: with audio, accuracy reward converges +0.1 higher than video-only training (Figure~\ref{fig:rl_curve1}, right), highlighting the importance of audio for video learning.

\begin{figure*}[t]
\centering
% \vspace{-30pt}
\includegraphics[width=\linewidth]{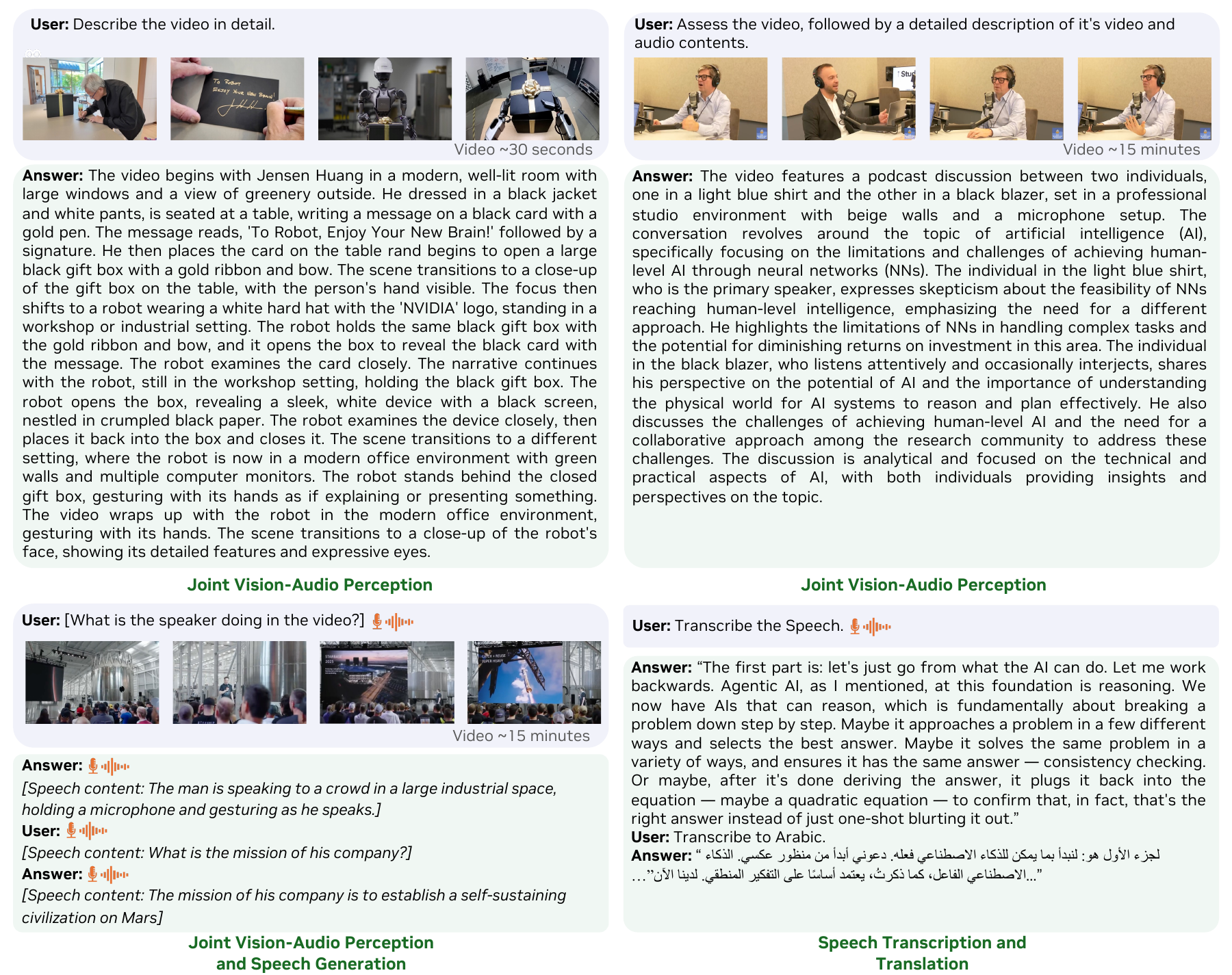}
\caption{\modelname demonstrates strong vision and audio perception capabilities to handle single or joint modality scenarios. The model also supports audio prompts and outputs. 
}
\label{fig:conv_video}
\end{figure*}

\begin{tcolorbox}[highlightstyle]
\keyinsight{} 
Joint audio-visual input surpasses the visual-alone input for GRPO training, offering faster and better convergence.
\end{tcolorbox}

\subsection{Downstream Tasks}
\modelname also improves downstream tasks that benefit from video-audio perception, including speech prompted robot navigation (Appendix Sec.~\ref{downstream:robot}),  sports video understanding (Appendix Sec.~\ref{downstream:sport}), cross-lingual speech translation (Appendix Sec.~\ref{downstream:speech}), medical analysis considering physician verbal explanations (Appendix Sec.~\ref{downstream:medical}), and semiconductor factory monitoring (Appendix Sec.~\ref{downstream:factory}). \modelname enables new frontier performances in these domains.

\subsection{Qualitative Study}
\label{app_sec:realworld}
To evaluate the performance of the model on real-world omni-modal signals, we test it using recently released online videos, as shown in Figure~\ref{fig:conv_video}. Our results demonstrate that the model can thoroughly comprehend both visual and audio inputs from previously unseen videos and generate responses based on this information, highlighting its strong generalization capabilities. 
The model successfully integrates speech cues with visual data, allowing for more effective interaction with the environment.
These qualitative observations demonstrate the effectiveness of the proposed explicit and implicit training strategy.

\section{Conclusion}
We present \modelname, a systematic effort to build an omni-modal LLM that allows joint perception of images, videos, audio, and text. We discuss architectural innovations including \alignmodelname, Temporal Embedding Grouping, and Constrained Rotary Time Embedding, joint with an enhanced data and training recipe. \modelname showcases frontier omni-modal performances, cuts down on training and inference costs, and improves downstream agentic applications.

\section{Acknowledgment}
We would like to express our gratitude to Tsung-Yi Lin, Yin Cui, and Ming-Yu Liu for their feedback on the system design. We also appreciate the discussions on data preparation with Wei Ping, Zhiding Yu, Guilin Liu, Shuoyang Ding, Zhehuai Chen, Jagadeesh Balam, Boris Ginsburg and Karan Sapra. Our study on medical AI data benefited from the support of Andriy Myronenko and Baris Turkbey. We also thank Shizhe Diao, Greg Heinrich, Yonggan Fu, Xin Dong, Peter Belcak, Baifeng Shi, Orr Zohar, Pritam Biswas, Boyi Li, Zhifeng Kong, Shinji Watanabe, and Tuomas Rintamaki for their helpful discussions. The audio encoders pre-training and speech post-training have been also supported by NVIDIA Taiwan Research and Development Center (TRDC) Program in 2025. We thank Eric Kang, Mason Wu, and Frank Lin for the NVIDIA DGX H100 infrastructure, partially used in this work.

\bibliography{custom}
\bibliographystyle{iclr2026_conference}

\clearpage
\appendix

\section*{Appendix}

\section*{Appendix Table of Contents}
\startcontents[appendix] %
\printcontents[appendix]{l}{1}{\setcounter{tocdepth}{3}}

\section{Related Works}
\label{app_sec:related}
A significant body of work has focused on augmenting LLMs with individual sensory capabilities, primarily vision and audio, often following a similar architectural blueprint. In the visual domain, the dominant paradigm involves using a vision encoder (\textit{e.g}., ViT~\citep{dosovitskiy2020vit}) to extract features which are then aligned with the LLM's input space via a bridging module. Pioneering models like Flamingo~\citep{alayrac2022flamingo} introduced sophisticated cross-attention mechanisms, while subsequent works~\citep{li2023blip,zhu2023minigpt,ye2023mplug,Driess2023PaLMEAE,liu2023llava,lin2023vila,liu2025nvila,mckinzie2024mm1,instructblip,damonlpsg2023videollama,wang2024internvideo2,Maaz2023VideoChatGPT,fang2024vila2,shi2025scaling,chen2025eagle2.5}, demonstrated the remarkable effectiveness of a simple projection layer combined with visual instruction tuning. A parallel line of research has applied this pattern to the auditory domain, where Audio-Language Models like LTU~\citep{gong2023listen}, Whispering-LLaMA~\citep{radhakrishnan2023whispering}, Audio-Flamingo~\citep{goel2025audio}, Qwen-Audio~\citep{chu2023qwen}, and others~\citep{tang2023salmonn,deshmukh2023pengi,kong2024audio,ghosh2025audio,huang2024audiogpt,chu2024qwen2} use audio encoders to process speech, music, and ambient sounds. These specialized models represent crucial stepping stones toward the more holistic goal of unified, omni-modal understanding.While specialized models for vision and audio have become increasingly capable, the development of foundational, omni-modal LLMs remains relatively nascent. For example, such a single omni model that can natively process and reason across text, vision, audio, and potentially other data types. 

The endeavor presents various challenges in terms of model architecture, data curation, and the immense computational resources required for training.
Recent pioneering efforts have addressed the challenges of multimodal understanding and reasoning. Google's Gemini~\citep{google2023gemini} represents a significant advancement as a natively multimodal model designed to seamlessly integrate and reason across interleaved text, images, audio, and video inputs. 
However, it remains proprietary and is not available to the open-source community.
Within the open-source community, several noteworthy efforts on omni-modal LLMs have been introduced~\citep{li2025baichuan,lu2024unified,wu2024next,ye2024x,chen2023vast,hu2025investigating,chen2023valor,liu2025ola,fu2024vita}, demonstrating strong capabilities in joint vision–audio understanding tasks. Among these, Phi-4-MM~\citep{abouelenin2025phi} and Qwen2.5-Omni~\citep{xu2025qwen25omni} achieve the strongest results to date; however, their accompanying technical reports reveal relatively simple architectural choices and a lack of thorough ablation studies to systematically examine critical design decisions.
In contrast, our work not only proposes several novel techniques for omni-modal understanding but also adopts a more rigorous experimental approach by conducting comprehensive ablation studies before scaling to large-scale datasets. We systematically evaluate various architectural choices and design decisions, providing detailed experimental analyses that we make publicly available. Through this methodical investigation, we aim to contribute valuable insights that can inform and inspire future research directions in omni-modal large language models.

\noindent\textbf{Compared Models in Experiments.}
In the experimental section, we compare with prior works on vision LLMs, audio LLMs, and omni-modal LLMs on various multimodal benchmarks. Specifically, we list the reference here due to the space limit in main text. Compared models include Gemini \citep{google2023gemini,google2024gemini}, GPT-4o \citep{openai2024gpt4o}, Claude-3.0 \citep{anthropic2024claude3}, InternVL2 \citep{chen2024internvl15}, Qwen2-VL \citep{wang2024qwen2vl}, Qwen2.5-Omni \citep{xu2025qwen25omni}, Phi-4-MM \citep{abouelenin2025phi}, Kimi-Audio \citep{kimiteam2025kimiaudiotechnicalreport}, Audio Flamingo 2 \citep{goel2025audio}, Qwen2 Audio~\citep{chu2023qwen}, Gemma~\citep{team2024gemma}, LTU~\citep{gong2023listen}, SALAMONN~\citep{tang2023salmonn}, Baichuan-Omni-1.5~\citep{li2025baichuan}, Whisper-large-3 \citep{radford2023robust}, LLaVA-NeXT-Video~\citep{zhang2024llavanext-video}, InternVL2~\citep{chen2024internvl15}, LLaVA-OneVision \citep{li2024llavaonevision}, LongVILA \citep{xue2024longvila}, Qwen2.5-VL \citep{wang2024qwen2vl}, NVILA \citep{liu2025nvila}, Video-ChatGPT~\citep{Maaz2023VideoChatGPT}, VideoChat2~\citep{li2024mvbench}.

\section{Downstream Agents}
\label{app_sec:agents}

Next, we demonstrate the applicability of \modelname in a wide range of downstream agentic tasks that yield consistent improvements across benchmarks while enabling new capabilities.

\begin{figure*}[t]
\centering
\includegraphics[width=.9\linewidth]{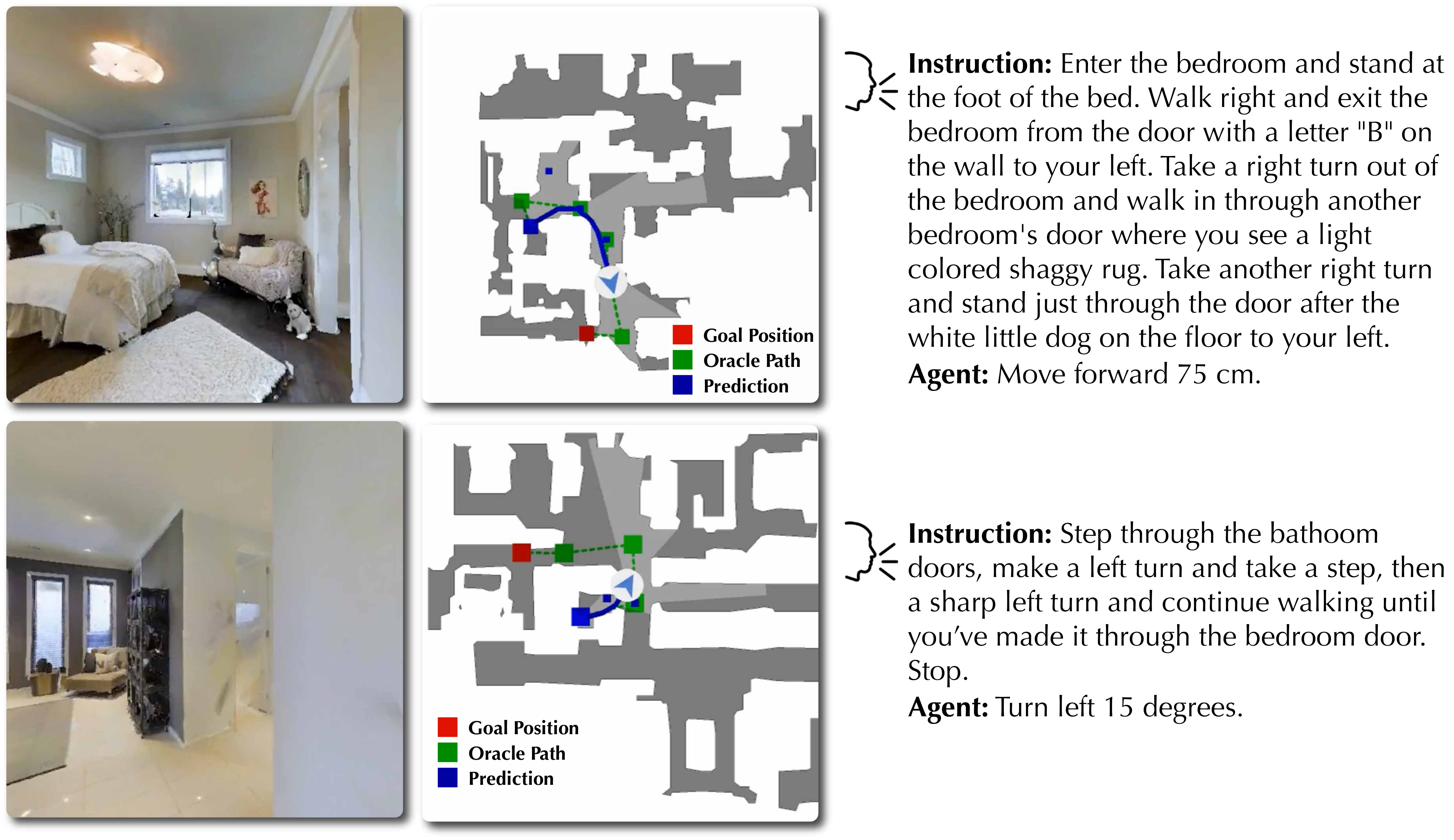}
\caption{An illustration of our speech-driven navigation agent based on \modelname. \textbf{Left:} Agent's current visual observation. \textbf{Middle:} Top-down map indicating the goal position and the agent's past trajectory. \textbf{Right:} the input speech instruction and the agent's predicted action given the current observation.}
\label{fig:demo_navila_r2r}
\end{figure*}

\subsection{Robotics: Speech-Driven Vision Language Navigation}
\label{downstream:robot}

Prior work~\citep{cheng2024navila,zhang2024navid,chen2023a2nav} in Vision-Language Navigation~\citep{anderson2018vision} has predominantly relied on text-based prompts. However, this is not always practical for real-world scenarios where the most convenient and natural way to command a robot is through human speech. As a first step toward this goal, we introduce a speech-driven vision language navigation task. This task is inherently more challenging than its text-based counterpart, as interpreting the nuances of speech is more complex than processing clean text.

\begin{table}[h]
    \centering
    \caption{Vision Language navigation results on R2R-CE. Our speech-driven model, \modelname, achieves comparable performance to the text-driven NVILA, with a lower navigation error.}
    \vspace{-10pt}
  \tablestyle{8pt}{1.2}
    \scriptsize
    \begin{tabular}{lcccccccc}
        \toprule
         \multicolumn{1}{c}{\multirow{2}{*}{Model}} & \multirow{2}{*}{Size} & \multirow{2}{*}{Obs.} & \multirow{2}{*}{Instruction} & \multicolumn{4}{c}{R2R Val-Unseen} \\
        \cmidrule(lr){5-8}
        & & & & NE $\downarrow$ & OS $\uparrow$ & SR $\uparrow$ & SPL $\uparrow$ \\
        \midrule
        Seq2Seq & -- & RGB & Text & 10.10 & 8.0 & 0.0 & 0.0 \\
        CMA & -- & RGB & Text & 9.55 & 10.0 & 5.0 & 4.0 \\
        NaVid & 7B & RGB & Text & 5.47 & 49.0 & 37.0 & 35.0 \\
        NVILA & 8B & RGB & Text & \textbf{5.43} & {60.4} & \textbf{53.3} & \textbf{48.8} \\
        \midrule
        \rowcolor{lightgreen}
        \modelname & 9B & RGB & Audio and/or Text & 5.67 & \textbf{60.8} & 50.6 & 45.1 \\
        \bottomrule
    \end{tabular}
    \label{tab:app:nav}
\end{table}

Specifically, we fine-tune \modelname on the training split of R2R-CE~\citep{r2r}, a benchmark for Vision-and-Language Navigation in continuous environments, with speech prompts, using 8 history frames for context in line with NVILA~\citep{liu2025nvila}.
As shown in the results in Table~\ref{tab:app:nav}, \modelname surpasses many text-based models and achieves performance comparable to NVILA.
We present qualitative examples in Figure~\ref{fig:demo_navila_r2r} that illustrate how our speech-driven vision-language-action (VLA) navigation agent functions in practice. The agent is deployed in the Habitat simulator under the continuous environment setting. The demo provides three synchronized views: (1) the agent’s current observation in RGB (left), (2) a top-down map indicating the goal location and the trajectory taken so far (middle), and (3) the spoken instruction together with the agent’s predicted action, such as moving forward a certain distance or turning left or right by a specified angle (right).

\subsection{Sport Video Understanding}
\label{downstream:sport}
Understanding videos of complex sports scenarios requires models to capture both visual dynamics and contextual cues. To evaluate the sports understanding capability of our proposed \modelname, we conduct experiments on the SPORTU-video dataset \citep{xia2025sportu}, a large-scale benchmark for fine-grained sports comprehension. As shown in Table~\ref{tab:sportu_results_xy}, \modelname-9B delivers strong performance despite its compact scale of 9 billion parameters. These results confirm the effectiveness of our model design and motivate its extension to more demanding, real-world applications such as live sports broadcasting, where both accuracy and efficiency are essential.

\begin{figure}[t]
\centering
\includegraphics[width=\linewidth]{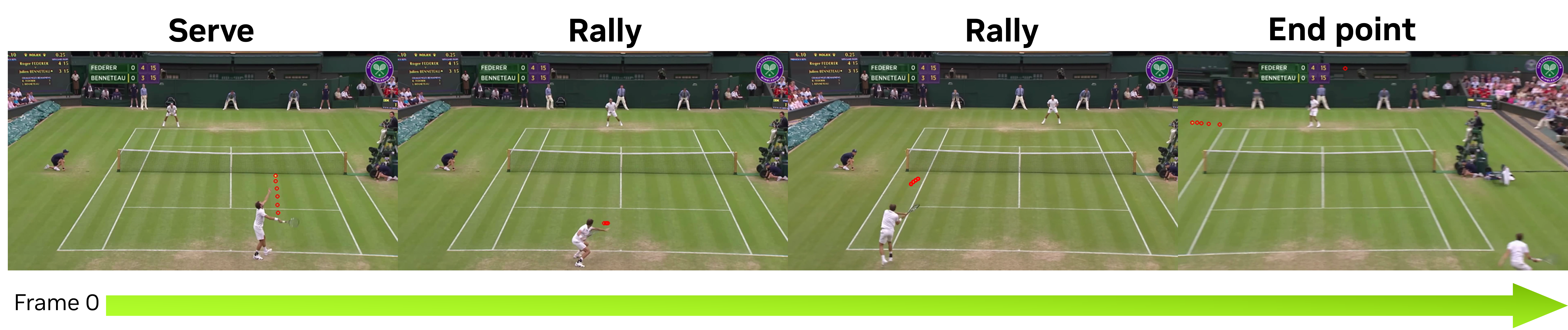}
\begin{minipage}{1\linewidth}
\scriptsize            
\raggedright
\vspace{0.6em}

\textbf{System Prompt:} Federer -- Roger Federer is the player in the white polo T-shirt and matching shorts. 
He has a white bandanna tied around his forehead and a white wristband on his right wrist. 
He is also wearing a pair of white shoes with white socks. 
Benneteau -- Julien Benneteau is the player in the white polo T-shirt with black accents and matching shorts. 
He has a white wristband on his right wrist and a white pair of shoes with black markings on them. 
He has a black ankle brace on his right foot as well.

\vspace{0.6em}
\textbf{Instruction:} What happened in this point? Provide a detailed caption.\\
\textbf{Agent:} Benneteau's first serve lands outside the service court, but his second attempt is successful. 
Federer wins the point with a lengthy rally where Benneteau hits the ball out of bounds with a forehand shot.

\vspace{0.6em}
\textbf{Instruction:} How this point ended between Federer and Benneteau.\\
\textbf{Agent:} The point ended when Benneteau hit the ball out of bounds.

\end{minipage}

\caption{Example of tennis broadcast commentary generation. 
For better visualization, we added red circle highlights to the tennis ball.
}
\label{fig:tennis_broadcast}
\end{figure}

\begin{table}[t]
  \centering
  \caption{Overall performance of MLLMs on SPORTU-video for multiple-choice questions. The best results within each category are \textbf{bolded}. Notably, our \modelname model 
  achieves highly competitive accuracy compared with both closed-source and open-source models.}
  \label{tab:sportu_results_xy}
  \vspace{-10pt}
  \tablestyle{12pt}{1.2}
  \scriptsize
  \begin{tabular}{lc}
    \toprule
    \textbf{Model} & \textbf{(Acc., $\uparrow$)} \\
    \midrule
    \multicolumn{2}{c}{\textit{Close-Source Model}} \\
    \midrule
    Claude-3.0-Haiku & 47.95 \\
    Gemini 1.5 Pro & 64.93 \\
    Gemini 1.5 Flash & 62.52 \\
    GPT-4omini & 58.19 \\
    GPT-4o & \textbf{68.79} \\
    \midrule
    \multicolumn{2}{c}{\textit{Open-Source Model}} \\
    \midrule
    ChatUniVi & 41.89 \\
    LLaVA-NeXT & 63.72 \\
    mPLUG-Owl3 & 60.80 \\
    ST-LLM & 46.39 \\
    Tarsier & 60.99 \\
    Video-ChatGPT & 34.05 \\
    VideoChat2 & 61.53 \\
    Qwen2.5-Omni-7B & 60.49 \\
    \modelname-9B (ours) & \textbf{67.30} \\
    \bottomrule
  \end{tabular}
\end{table}

To further assess performance in realistic broadcasting settings, we curate a tennis-specific dataset collected from 14 full matches. The dataset contains 24,078 multiple-choice questions and 20,214 open-ended questions derived from pre-clipped videos, each spanning 3–120 seconds with precisely annotated start and end points. Since sports broadcasting requires synchronizing visual actions with speech cues (\textit{e.g.}, live commentators’ narration or umpire calls) to enable professional-style commentary, tennis provides an ideal domain for multimodal evaluation.

In our tennis experiments, we evaluate tasks such as identifying the server from player characteristics, determining the point winner, and classifying the outcome type (\textit{e.g.}, ace, forced error, unforced error). The benchmark \modelname processes clips at their native resolution (primarily FHD $1920 \times 1080$), using 128-frame segments per point. As shown in Table~\ref{tab:tennis_results}, \modelname substantially outperforms Qwen2.5-Omni in predicting point outcomes and rally length, demonstrating the advantages of high-resolution spatiotemporal modeling. Figure~\ref{fig:tennis_broadcast} illustrates sample videos with action explanations, along with generated open-ended commentary styled after professional broadcasters.

For efficient deployment, we adopt the LLM-AWQ implementation of Activation-aware Weight Quantization~\citep{lin2024awq}, which enables 4-bit quantization while preserving accuracy. Inference is executed using the TinyChat engine on NVIDIA hardware, supporting multimodal video–audio inputs. On a single NVIDIA A100, \modelname achieves an average latency of under 2 seconds per pre-quantized clip, delivering a 45\% boost in inference speed and making it well-suited for live broadcasting scenarios. We further validate deployment on NVIDIA L40s GPUs, demonstrating the practicality of our approach in resource-constrained environments.

\begin{table}[t]
  \centering
  \caption{Comparison of video understanding accuracy (\%) for tennis broadcasting. Results are evaluated with multiple-choice questions (MCQ). Inference time is measured on an NVIDIA A100, with input clips averaging around 20 seconds in duration. AWQ indicates model quantization performed with the AWQ technique \citep{lin2024awq}.}
  \label{tab:tennis_results}
    \vspace{-10pt}
  \tablestyle{8pt}{1.2}
  \scriptsize
  \begin{tabular}{l c c c c c}
    \toprule
    Model & \makecell{Inference Time \\ (Seconds $\downarrow$)} & \makecell{Server \& \\ Winner} & \makecell{Receiver \& \\ Winner} & \makecell{Point \\ Ending} & \makecell{Shots \\ Exchanged} \\
    \midrule
    Qwen2.5-Omni & 3.34 & 96.2 & 90.7 & 48.6 & 38.3 \\
    \textbf{\modelname}       & 3.29 & \textbf{100.0} & \textbf{100.0} & \textbf{85.7} & \textbf{89.3} \\
   \textbf{\modelname w/ AWQ} & \textbf{1.85} & \textbf{100.0} & \textbf{100.0} & \textbf{85.7} & 85.1 \\
    \bottomrule
  \end{tabular}
\end{table}

\subsection{Speech Agent: Speech Translation} 
\label{downstream:speech}
\begin{table}[t]
  \centering
  \caption{Performance comparison of different models on Covost2 speech translation tasks measured by BLEU scores. EN $\rightarrow$ X denotes translation from English to the target language, and X $\rightarrow$ EN denotes translation from the target language to English. Languages: zh = Chinese, ja = Japanese, ar = Arabic, de = German. }
  \vspace{-10pt}
  \tablestyle{8pt}{1.2}
  \scriptsize
  \begin{tabular}{lccccc|ccccc}
    \toprule
    \multicolumn{1}{c}{\multirow{2}{*}{\textbf{Model}}}  & \multicolumn{5}{c|}{\textbf{EN $\rightarrow$ X} (Acc., $\uparrow$)} & \multicolumn{5}{c}{\textbf{X $\rightarrow$ EN} (Acc., $\uparrow$)} \\
    & {zh} & {ja} & {ar} & {de} & {avg.} & {zh} & {ja} & {ar} & {de} & {avg.} \\
    \hline
    Qwen2-audio   & \textbf{45.2} & 24.8 & 20.1 & 29.9 & 30.0 & 24.4 & 18.7 &  19.5 & 35.2 & 24.5 \\
    Qwen2.5-omni & 41.4 & 26.0 & 19.7 & 30.2 & 29.3  &29.4 & 12.1 & 19.3 & 37.7 & 24.6 \\
    Phi-4-mm      & 38.0 & {31.9} & 9.9  & 35.3 & 28.9 & 24.9 &  33.3 & 5.5  & \textbf{37.9} &25.7 \\
    \rowcolor{lightgreen}
    \textbf{\modelname} & 39.7 & \textbf{32.6} & \textbf{23.3} & \textbf{35.5} & \textbf{32.8} &  \textbf{29.9} & \textbf{33.7} & \textbf{20.1} & 32.6 & \textbf{29.1}\\ %
    \bottomrule
  \end{tabular}
  \label{tab:speech_translation}
\end{table}

We benchmark \modelname on the CoVoST2~\citep{wang2020covost} speech translation task, measuring BLEU scores across multiple target languages in both EN~$\rightarrow$X and X$\rightarrow$EN directions, after fine-tuning on related data, and show the results in Table~\ref{tab:speech_translation}. 
Our model delivers competitive translation quality across most directions, with particularly strong performance in X~$\rightarrow$~EN for Japanese ({23.2 BLEU}) and Arabic ({23.0 BLEU}). This balance of accuracy across languages highlights the benefit of integrating speech translation corpora within our omni-modal training pipeline, enabling  to perform both recognition and translation in a unified framework. The ability to handle multilingual speech understanding and cross-lingual transfer further broadens the applicability of our model in real-world communication, international dialogue systems, and cross-border information access.

\subsection{Medical AI}
\label{downstream:medical}

\begin{figure*}[t]
\centering
\includegraphics[width=0.7\linewidth]{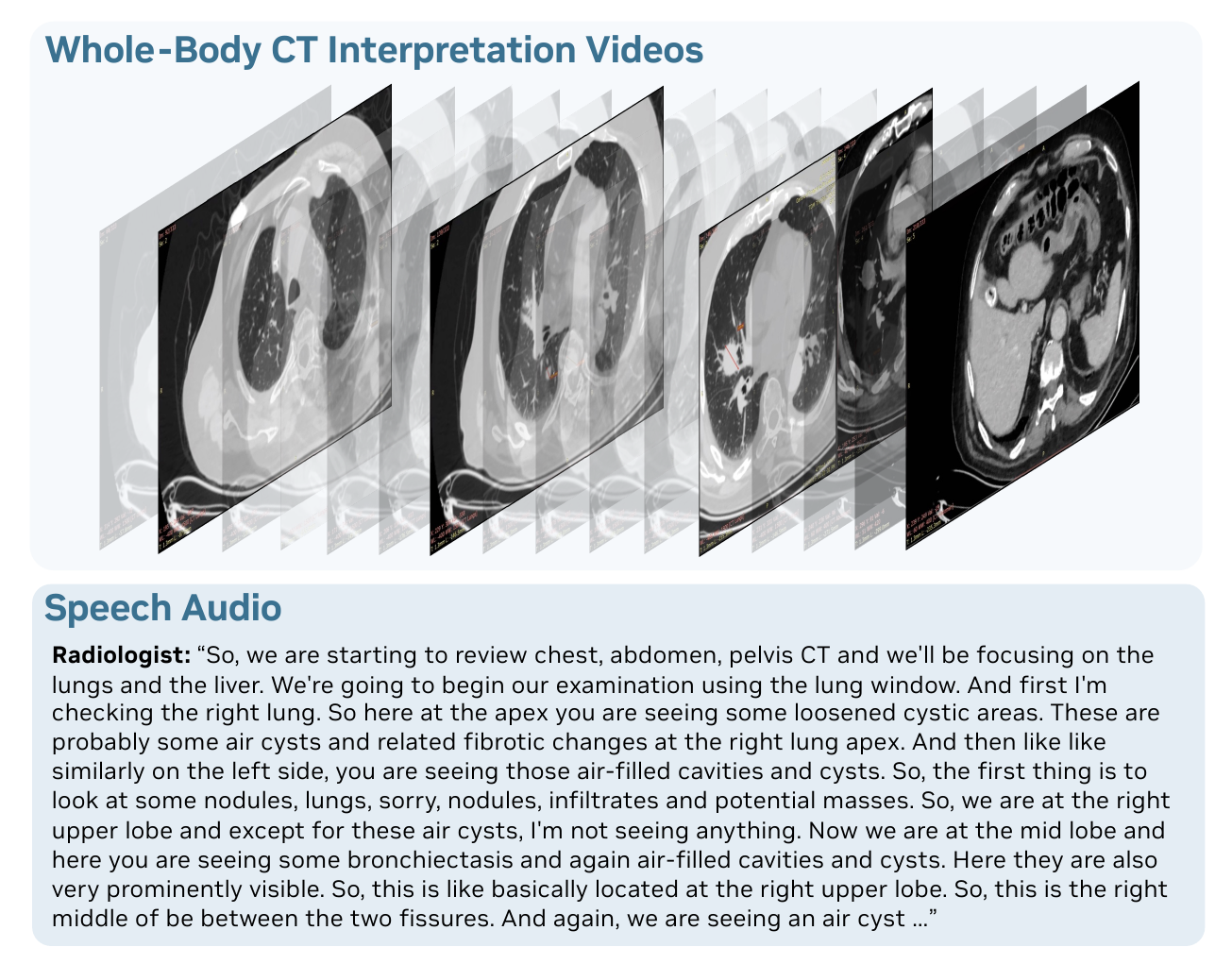}
\vspace{-10pt}
\caption{Sample frames and transcript trunks from one of the curated radiologist-narrated CT interpretation video. For annotation, the radiologist maintains a 2D axial view while progressively adjusting visualization (\textit{e.g.}, window/level, zoom) and annotating across slices.}
\label{fig:med-video-audio}
\end{figure*}

We evaluate \modelname's zero-shot generalization to the medical domain using 49 privacy-deidentified, radiologist-curated video clips of whole-body CT interpretations.
As illustrated in Figure~\ref{fig:med-video-audio}, each 2-minute recording captures a radiologist interpreting real-world clinical images with a 2D axial-plane viewer, including scrolling through slices, placing measurements and annotations, zooming, adjusting window/level, and, when relevant, comparing the same image under different window settings.

From these video–audio pairs and their transcripts, we construct 588 multiple-choice questions spanning four categories—(i) long-horizon temporal reasoning and localization, (ii) audio–visual synchronization and understanding, (iii) anti-shortcutting (resisting language priors without visual evidence), and (iv) temporal reasoning—approximately balanced across categories with three options per item.
The dataset was curated with assistance from the LLama-3.1-Nemotron-Ultra-253B~\citep{bercovich2025llamanemotronefficientreasoningmodels}, leveraging both the visual content and transcripts. 
We report comparative performance for \modelname and Qwen2.5-Omni in Table~\ref{tab:med_video_results}.

\begin{table}[h]
\centering
\caption{Performance comparison between \modelname and Qwen2.5-Omni on omni-modal multiple-choice QA datasets across four categories.
Abbreviations: LH = long-horizon temporal reasoning \& localization; AVS = audio-visual synchronization \& understanding; AS = anti-shortcutting (resisting language priors without video evidence); TR = temporal reasoning.}
\label{tab:med_video_results}
    \vspace{-10pt}
  \tablestyle{8pt}{1.2}
  \scriptsize
\begin{tabular}{lccccc}
\toprule
\multicolumn{1}{c}{Method} & Acc.~(LH) $\uparrow$ & Acc.~(AVS) $\uparrow$ & Acc.~(AS) $\uparrow$ & Acc.~(TR) $\uparrow$ & Average $\uparrow$\\
\midrule
Qwen2.5-Omni               & 0.83 & 0.75 & 0.91 & 0.70 & 0.79  \\
\rowcolor{lightgreen}
\textbf{\modelname}                  & \textbf{0.84 }&\textbf{ 0.76} & \textbf{0.92 }& \textbf{0.76} &\textbf{ 0.82} \\
\bottomrule
\end{tabular}
\end{table}

\begin{figure*}[t]
\centering
\includegraphics[width=1\linewidth]{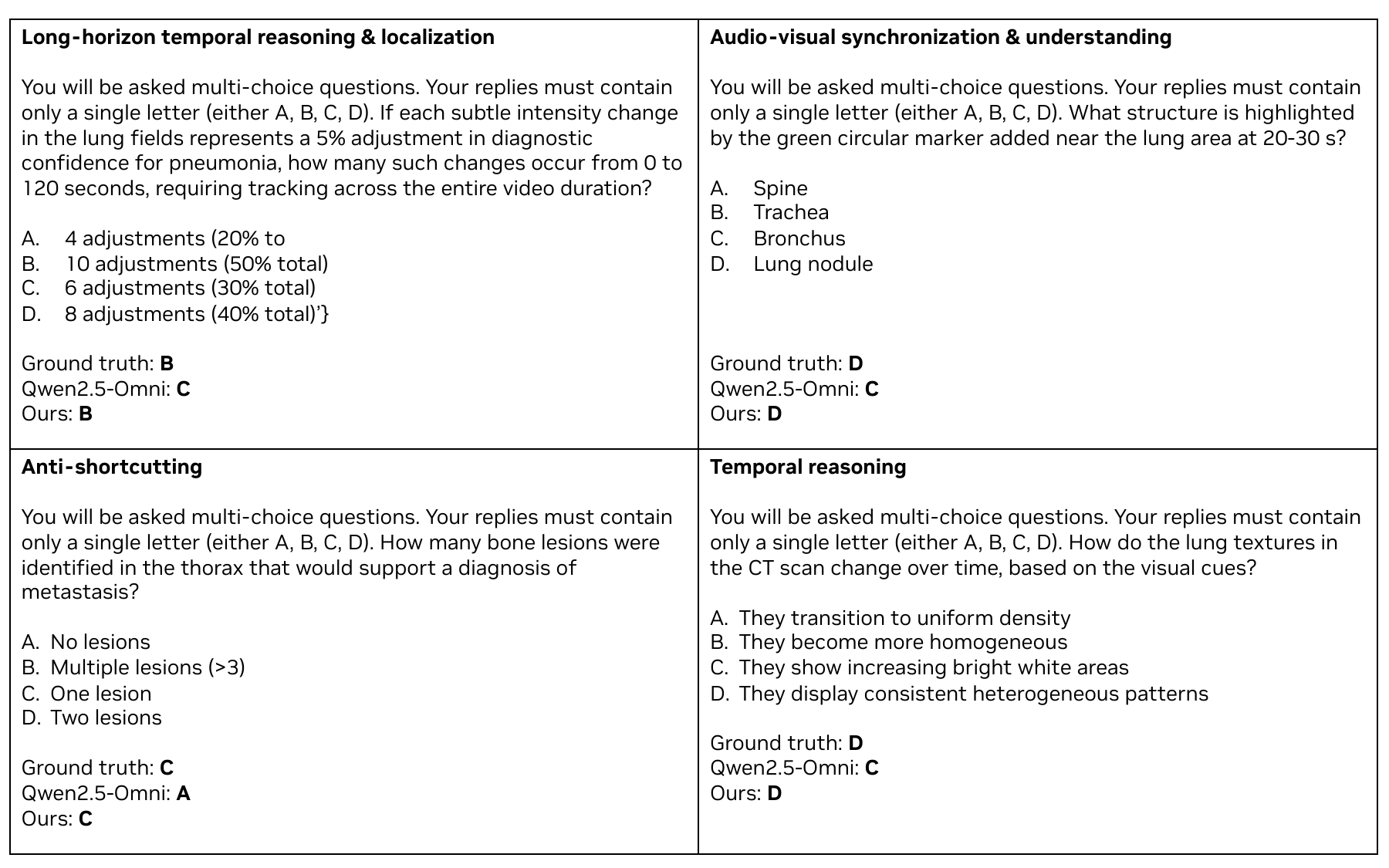}
    \vspace{-10pt}
\caption{Qualitative comparison between \modelname and Qwen2.5-Omni on an omni-modal medical QA task based on radiologist-narrated CT interpretation videos. We organize the evaluation into four categories of questions: long-horizon temporal reasoning and localization, audio-visual synchronization and understanding, anti-shortcutting, and temporal reasoning.
}
\label{fig:med-qualitative}
\end{figure*}

\modelname consistently outperformed Qwen2.5-Omni across all four categories, yielding an overall gain of about +2.0 percentage points. Its largest margin was in temporal reasoning (TR; +6.1), highlighting stronger capabilities in event sequencing, change detection, and temporal cue modeling—often the most demanding aspects of video comprehension in clinical workflows. Stable improvements were observed in long-horizon reasoning (LH) and audio-visual synchronization (AVS) (+0.7 each), reflecting better preservation of long-range context and closer alignment between narration and visual content. The anti-shortcutting (AS) category also showed a gain of +0.7, suggesting that \modelname is more robust against linguistic shortcuts and leans more heavily on visual evidence. 
Some qualitative comparisons of test samples are presented in Figure~\ref{fig:med-qualitative}.

\subsection{Smart Factory Agents} 
\subsubsection{Semiconductor Manufacturing} 

\label{downstream:factory}

\begin{figure*}[h!]
\centering
\includegraphics[width=1\linewidth]{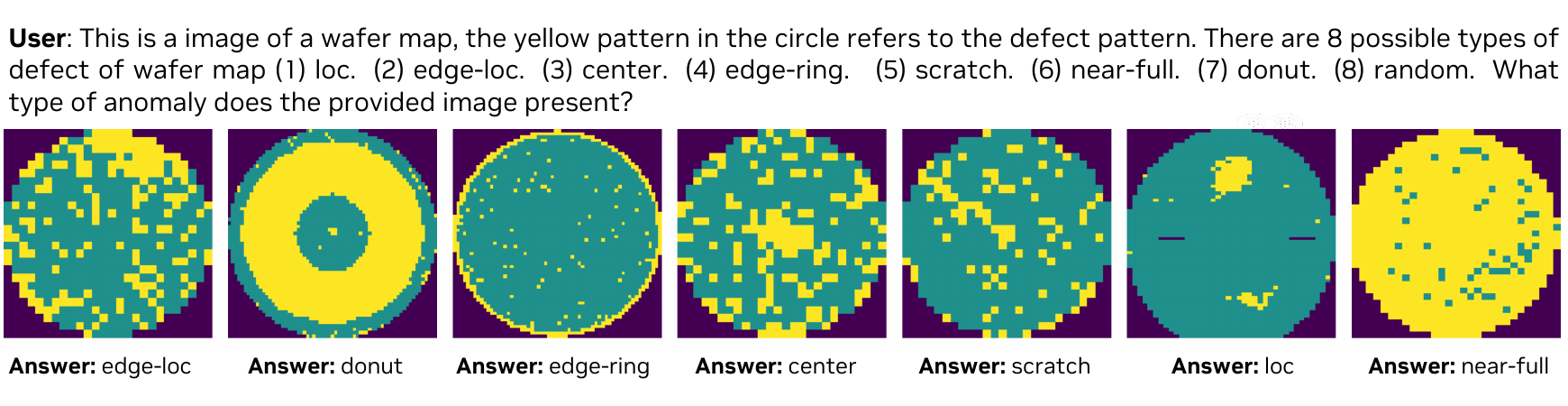}
    \vspace{-10pt}
\caption{Illustration of wafer robust defect analysis task for smart factory agent. 
}
\label{fig:wafer_overview}
\end{figure*}

Wafer maps are essential in semiconductor manufacturing for visualizing defect distributions, enabling yield monitoring, process drift detection, and preliminary root cause identification. 
It is a domain with a significant gap from multimodal LLM.
To study whether we can leverage our omni-modal \modelname on this task, we fine-tune \modelname on wafer map data, aligning visual and textual features for robust defect analysis, as illustrated in Figure~\ref{fig:wafer_overview}. 
On the WM-811K dataset~\citep{wu2015wafer}, \modelname achieves superior performance over VILA~\citep{lin2023vila} and NVILA~\citep{lin2025genai,liu2025nvila} (which has been trained for wafer defect classification), and our model demonstrates further improvements, as summarized in Table~\ref{tab:wafer_results}. Beyond classification, this framework can be extended to support interactive querying and automated reasoning for Root Cause Analysis, systematically linking defect clusters to process tools, wafer locations, or temporal drifts.

\begin{table}[h]
  \centering
  \caption{Comparison of VILA, NVILA, and \modelname on wafer defect classification. 
  }
    \vspace{-10pt}
  \tablestyle{8pt}{1.2}
  \scriptsize
  \begin{tabular}{lccc}
    \toprule
     & \textbf{VILA}~\citep{lin2023vila} & \textbf{NVILA}~\citep{liu2025nvila} & \textbf{\modelname} (ours) \\
    \midrule
    Parameters & 40B & 8B  & 9B \\
    Resolution & 336$\times$336 & 448$\times$448 & 448$\times$448 \\
    Model size & 75 GB & 16 GB & 18 GB\\
    Accuracy   & 90.8\% & 97.6\% & \textbf{98.1\%} \\
    \bottomrule
  \end{tabular}
  \label{tab:wafer_results}
\end{table}

\begin{figure*}[h]
\centering
\includegraphics[width=.8\linewidth]{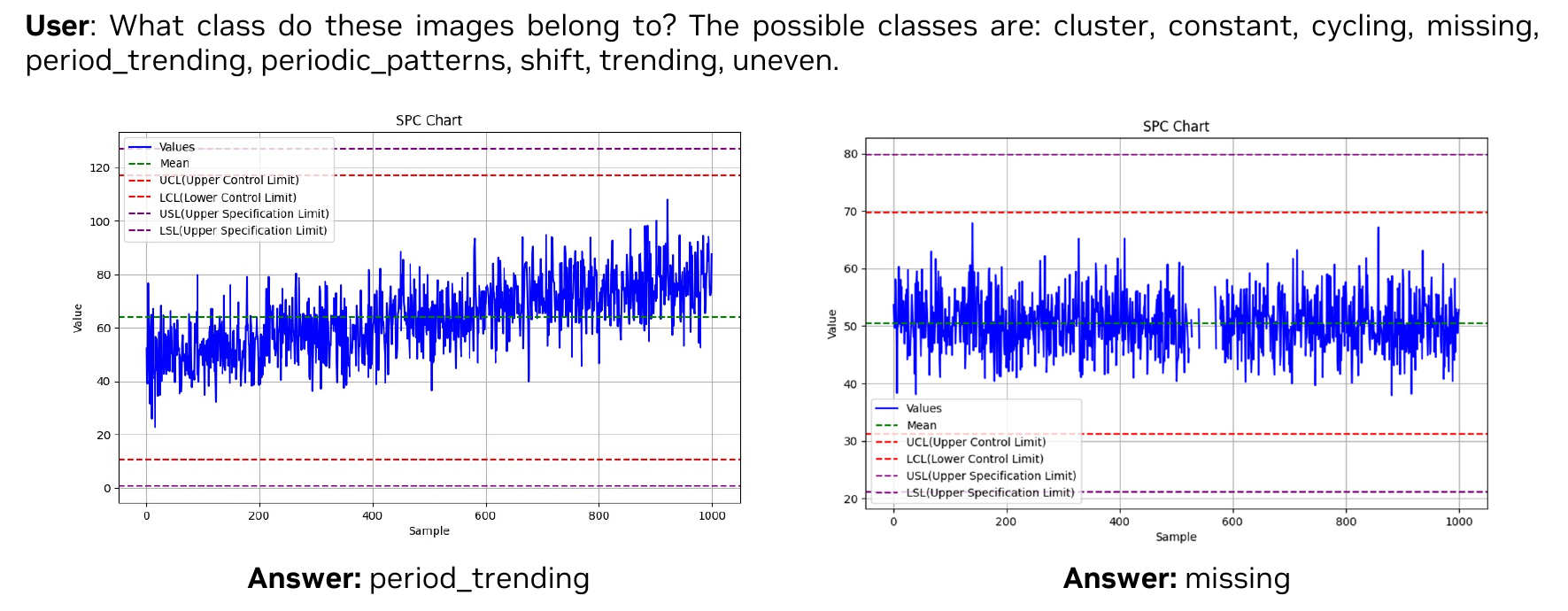}
\caption{Illustration of SPC chart recognition for industrial fault detection.}
\label{fig:SPC_Charts}
\end{figure*}

\subsubsection{Factory and Industrial Time Series Understanding}
We apply \modelname to Statistical Process Control (SPC) chart recognition, a representative task in industrial quality monitoring and root cause analysis. Our model recognizes a wide range of fault categories, including out-of-control points such as spikes or drops, persistent runs and monotonic trends such as level shifts up or down, cyclic oscillations, mixture or random fluctuations, as well as missing values or short outages, as illustrated in Figure~\ref{fig:SPC_Charts}. On a held-out test set, our model achieves 87\% accuracy, showing that by transforming time-series signals into visual representations, we can effectively leverage large-scale vision-language pretraining for sensor monitoring and industrial diagnostics. This demonstrates the feasibility of deploying our framework in real manufacturing pipelines, where timely detection of process abnormalities is crucial for preventing defects and reducing downtime.

We assess our framework on time series classification tasks using datasets from the UCR archive~\citep{dau2018ucr}, where time series are transformed into line plots to exploit large-scale vision–language pretraining. Our first comparison is against VLM-TSC~\citep{prithyani2025VLMTSC}, a LLaVA-based VLM that adopts a similar conversion strategy. As shown in Table~\ref{tab:ucr_results_vlm_compact}, our approach achieves superior performance on the PenDigits and ItalyPowerDemand datasets.

\begin{table}[t]
  \centering
  \caption{Performance comparison of test accuracy (\%) on selected UCR datasets~\citep{dau2018ucr}.}
    \vspace{-10pt}
  \label{tab:ucr_results_vlm_compact}
  \tablestyle{8pt}{1.2}
  \scriptsize
  \begin{tabular}{l c cccc|cccc}
    \toprule
    \multicolumn{6}{c|}{\textbf{Dataset Info}} & \multicolumn{2}{c}{\textbf{Acc.} $\uparrow$} \\
    \cmidrule(lr){1-6} \cmidrule(lr){7-8}
    \textbf{Dataset} & \textbf{Type} & \textbf{Length} & \textbf{Train} & \textbf{Test} & \textbf{Class}
            & \textbf{VLM-TSC~\citep{prithyani2025VLMTSC}} & \textbf{Ours} \\
    \midrule
    PenDigits          & MOTION & 8   & 7494 & 3498 & 10 & 85.08 & \textbf{96.88} \\
    ItalyPowerDemand   & SENSOR    & 24  &   67 & 1029 &  2 & 95.00 & \textbf{95.82} \\
    \bottomrule
  \end{tabular}
\end{table}

\section{Method Details}
\label{app_sec:method}

\subsection{Omni-Modal Input Embedding}
\label{sec:omni_embed}
\noindent\textbf{Image.}
Similar to NVILA~\citep{liu2025nvila}, we start with pretrained the SigLip~\citep{zhai2023siglip} \footnote{Model version ``\texttt{paligemma-siglip-so400m-patch14-448}''} vision encoder and augment it with $2\times2$ ``Spatial Scale-Then-Compress'' Dynamic S2~\citep{liu2025nvila,shi2024we} to accomondate for multi-scale and high resolution images. Given an input image of varying dimensions, the overall encoding module adapts the largest scale to the nearest tile-aligned size divisible by 448 and maintains the aspect ratio. Feature maps from all scales are aligned to this largest scale and concatenated, processed by a 2-layer MLP for projection into a latent space shared by embeddings of different modalities.

\noindent\textbf{Audio.}
We adopt a single audio encoding pipeline for both speech and non-speech audio.
Raw audio waveforms are sampled at 16 kHz and converted into audio frames using the Short-Time Fourier Transform (STFT). These frames are then processed by the Audio Flamingo 3~\citep{goel2025audio} audio encoder to extract acoustic features in both speech and natural sound. The encoder consists of convolutional layers followed by transformers, enabling it to capture both local and global audio patterns. The extracted features are subsequently projected into the modality-shared latent space using a 2-layer MLP.

\noindent\textbf{Video.}
Videos contain two modalities introduced above, namely vision and audio. For the vision stream, the video frames are temporally sampled uniformly to reduce redundancy and computational load. Each frame is processed individually through the above-mentioned image input pipeline, and the resulting features are aggregated temporally. 
We then utlize temporal pooling on the feature sequence to further compress visual information. For {audio stream}, we extract features with the same Audio modality pipeline mentioned above. 
Meanwhile, we extract the timestamps for each visual and audio embeddings to act as temporal guidance on interleaved token arrangement as explained later.

\noindent\textbf{Prompt.}
For text prompts, we employ a standard text encoder~\citep{qwen25}, which first tokenizes the input into discrete tokens and then maps them into a continuous semantic embedding space via an embedding layer. This space is shared with embeddings from other modalities.
For speech prompts, we use the previously described audio encoder to generate the corresponding continuous semantic embeddings.
Finally, the resulting prompt embeddings are concatenated with the visual and audio embeddings introduced earlier.

\subsection{Modality-Specific Training}
\label{sec:modality_specific_train}

\subsubsection{Vision Training} 

The modality-specific vision training aims to train the model with visual understanding ability.
We follow NVILA~\citep{liu2025nvila} training recipe including five stages: 

\noindent\textbf{Stage 1 | Vision Projector Alignment.}
This stage learns to project visual information through a visual projector. This stage ensures that the visual embeddings are compatible with the language model’s token embeddings, which is essential for smooth downstream integration. The model is trained on image-text pairs with simple captioning-style supervision, setting a baseline understanding of visual semantics.
Only the vision projector is tuned during this process.

\noindent\textbf{Stage 2 | Vision Encoder Alignment.}
With the projector aligned, the model now focuses on enhancing the vision encoder’s capacity to process diverse visual content. In this stage we train only the vision encoder and visual projector.

\noindent\textbf{Stage 3 | Vision Pre-Training.}
During this core stage, the model is trained on large-scale multimodal data to learn how to interpret and generate image descriptions. The vision encoder is kept frozen, while the vision projector and the LLM are fine-tuned.

\noindent\textbf{Stage 4 | Image Instruction Tuning.}
In this stage the model is fine-tuned with vision instruction-following capabilities. It is trained to answer multimodal questions, generate captions, reason over scenes, interpret documents, and more. Training data covers a broad range of multimodal capabilities. It includes high-quality instructional examples to align the model with human preferences, datasets for generating rich image captions, and tasks that develop logical and visual reasoning skills. The model is also trained to interpret documents and embedded text, answer general and knowledge-based visual questions, and handle diagrams, visual dialogues, and multimodal instructions. 
In this stage, all model parameters are fine-tuned.

\noindent\textbf{Stage 5 | Video Instruction Tuning.}
In the final vision alignment stage, the model is adapted to video understanding.
The goal here is to enable temporal reasoning and visual understanding over sequences of frames. This includes tasks such as activity recognition, multi-frame object tracking, and answering time-sensitive questions.
The whole model is fine-tuned.

Through this vision alignment process, we obtain the ``vision preliminary checkpoint'' with well-trained vision encoder, projector, and language model.

\subsubsection{Audio Training} 
Starting from the language model in the above vision preliminary checkpoint we next train the audio understanding ability of our model, which involves (i) audio projector and encoder alignment step followed by (ii) audio instruction tuning.

\noindent\textbf{Stage 1 | Audio Projector \& Encoder Alignment.}
This phase focuses on aligning audio encoder and its associated compression layer. We keep the parameters of the language model and vision side fixed. Training consumes 50K audio-language pairs curated from public datasets spanning across audio-based (music, non-speech sound, and speech) question answering, speech-to-text captioning, and automatic speech recognition. By training on this heterogeneous dataset, we encourage the audio projection module to learn a unified representation that aligns well with the language model’s semantic space.

\noindent\textbf{Stage 2 | Audio Instruction Tuning.}
During the second stage of training, the audio encoder, audio projection module, and language model are fine-tuned in a unified, end-to-end manner. This joint optimization allows the system to develop a comprehensive and deeply integrated understanding of audio. This stage consumes a comprehensive audio-SFT dataset overseeing 9.6 million samples, including but not limited to audio-based question answering~(AudioEntailmentQA~\citep{deshmukh2025audio}, Clotho-AQA~\citep{lipping2022clotho}, DCASE-2025-train~\citep{Yang2025}, etc.), audio captioning~(AudioCaps~\citep{kim2019audiocaps}, Clotho-v2~\citep{drossos2020clotho}, Miradata~\citep{ju2024miradata}-recaptioned, etc.), speech emotion recognition (CREMA-D~\citep{cao2014crema}, IEMOCAP~\citep{busso2008iemocap}, MELD~\citep{poria2018meld}, etc.), automatic speech recognition (CV-ASR~\citep{commonvoice:2020}, Europarl-ASR~\citep{koehn2005europarl}, LibriSpeech-ASR~\citep{panayotov2015librispeech}, etc.), and speech translation (MuST-C~\citep{di2019must}, Emilia~\citep{he2024emilia}, etc.). This allows the model to learn both low-level acoustic features and high-level semantic representations, enabling robust generalization across multiple audio understanding tasks and versatile capabilities in interpreting complex auditory inputs. 
At this point, we find that the model's ability to perform visual understanding tasks is worse, which motivates us to pursue the subsequent omni-modal joint training.

\subsection{Omni-Modal Joint Training Details}
\label{sec:detail_joint_train}
We adopt a cosine learning rate schedule, preceded by a linear warm-up phase over the first 3\% of the training data. The base learning rate is set to $2\times10^{-5}$. During training, the vision and audio encoders are kept frozen.
The total token count is approximately 200 billion.

\subsection{Extra Details of Training Data}
\label{sec:detail_training_data}

This section describes the comprehensive multi-modality training data used for developing the proposed omni-modal LLM, which are designed to handle diverse types of audio, visual, and textual information. Our training corpus encompasses a wide range of modalities including speech recognition, audio question answering, audio captioning, audio classification, video question answering, and image understanding tasks. The dataset is carefully curated to provide robust coverage across multiple domains, enabling the model to develop strong cross-modal understanding and reasoning capabilities.

There are 3.6 million omni-modal conversations, 8 million image-text conversations, 2.7 million video-text conversations, 5.3M speech-text conversations, and 4.3 million speech-text conversations.  
Omni-modal data contributes 15\%, consisting of omni question answering (12\%) and omni captioning (3\%). Image data constitutes the largest share at 36\%, with notable subcategories including general image tasks (19\%), knowledge-based tasks (8\%), and document processing (7\%). Sound~(non-speech) data accounts for 21\%, predominantly driven by audio question answering (20\%). Speech data represents 17\% of the total, primarily comprising automatic speech recognition (10\%), audio question answering (4\%), and speech translation (2\%). Video data forms the remaining 11\%, entirely attributed to video question answering. 
The training data consists of approximately 24 million samples distributed across three main categories: Speech, Sound, and Image/Video. 
The Speech category includes datasets for automatic speech recognition (ASR), speech translation, and emotion classification, featuring well-established corpora such as AMI \citep{carletta2007unleashing}, Common Voice \citep{commonvoice:2020}, and LibriSpeech \citep{panayotov2015librispeech}. The Sound category encompasses audio question answering datasets like MMAUQA \citep{goel2025audio} and CompA-R-AQA \citep{ghoshcompa}, audio captioning datasets such as Clotho-v2 \citep{drossos2020clotho}, and various audio classification datasets including UrbanSound8K \citep{mesaros2018multi} and FSD50k \citep{fonseca2021fsd50k}. The Image/Video category includes datasets for visual question answering, document understanding, and general image/video comprehension tasks.
We include video training data from NVILA~\citep{liu2025nvila} and Eagle-2.5~\citep{chen2025eagle2.5}.

\section{More Experiments}
\label{app_sec:more_experiment}

\begin{figure*}[t]
\centering
\includegraphics[width=1\linewidth]{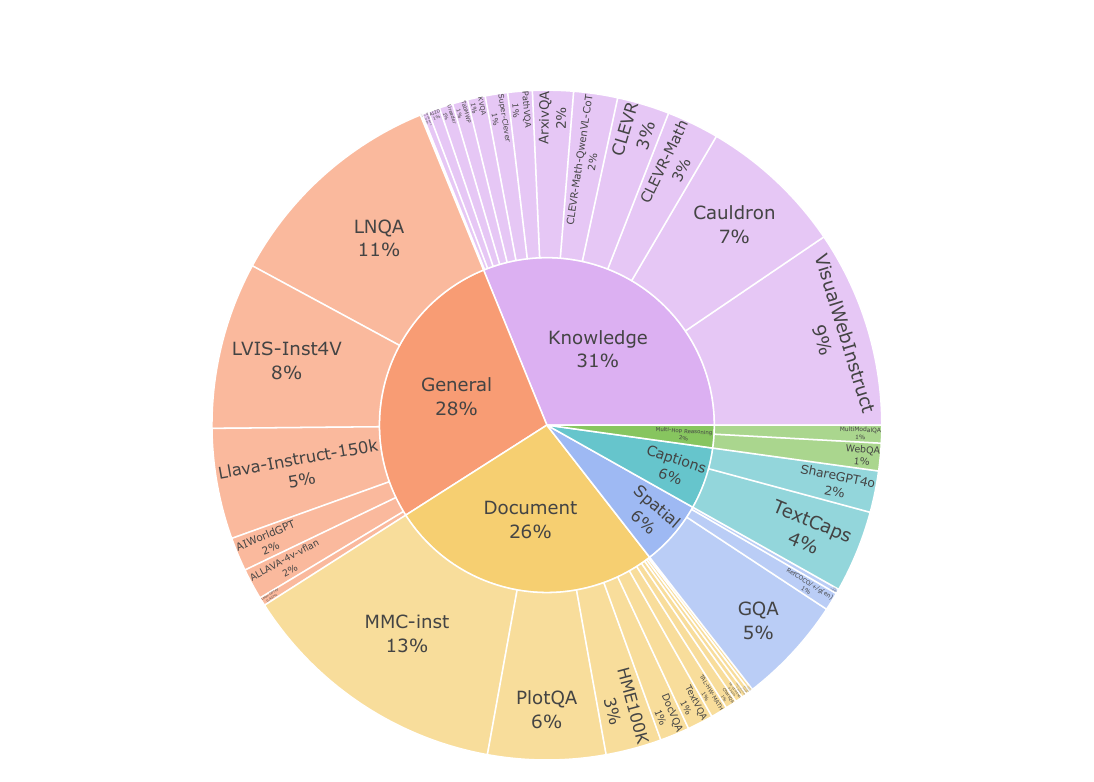}
\caption{Data distribution of our synthetic speech-prompted multimodal conversation.
}
\label{fig:speech_vqa_data}
\end{figure*}

\subsection{Audio Encoding}

\noindent
\textbf{Audio Encoder Backbone.} To investigate the choice of audio representations for the omni-modal model, we evaluate two state-of-the-art audio encoders: Qwen2-Audio~\citep{chu2023qwen} used by Qwen2.5-Omni~\citep{xu2025qwen25omni}, and the AF-Whisper backbone~\citep{goel2025audio} from Audio Flamingo 3~\citep{goel2025audio}. 
This comparative analysis enables us to identify the backbone that provides the most effective encoding for downstream multimodal tasks. 
Specifically, we ablate these key components by aligning them with the LLM backbone model we used in audio-only training.
We use 10\% of the audio/speech training data to fairly evaluate the effectiveness of the two encoders under the same data budget.
As shown in Table~\ref{tab:audio_encoder_ablation}, AF-Whisper consistently outperforms the Qwen-2 Audio encoder backbone on audio and speech understanding tasks. 
Therefore, our final model architecture adopts the AF-Whisper backbone to extract informative audio features.

\begin{table}[!h]
  \centering
  \caption{Ablation study on different Audio Encoder backbones.}
    \vspace{-10pt}
  \tablestyle{4pt}{1.2}
  \scriptsize
  \begin{tabular}{lcccc}
    \toprule
    \textbf{Audio Encoder} & \textbf{LS-clean} & \textbf{LS-other} & \textbf{MMAU-mini} & \textbf{MMAU}  \\
    \midrule
    Qwen2-Audio & 5.5 & 7.1 & 61.5 & 59.0\\
    \rowcolor{lightgreen}
    AF-Whisper -- \textbf{chosen} & \textbf{2.1} & \textbf{5.2} & \textbf{70.5} & \textbf{63.3} \\    
    \bottomrule
  \end{tabular}
  \label{tab:audio_encoder_ablation}
\end{table}

\noindent \textbf{Audio Token Compression.}
For the AF-Whisper encoder, similar to Whisper-large-v3~\citep{radford2023robust}, the process begins by resampling the audio to a 16 kHz sampling rate, followed by transforming the raw waveform into a 128-channel mel-spectrogram using a 25 ms analysis window and a 10 ms hop interval (\textit{i.e.}, a hop length of 160).
This yields 3,000 audio frames for a 30-second audio, which are then processed through convolutional layers and a transformer model to extract audio features, resulting in 750 sequential audio feature vectors. 
Therefore, each second of audio is roughly represented by 25 tokens.
While this may not seem like a lot for a 30-second audio, encoding one hour of audio would require about 90,000 tokens, which could overwhelm the context length of multimodal models.

We next explore several audio information compression strategies to improve efficiency in representing audio information. In our ablation study, we fine-tune the preliminary checkpoint before large-scale training on a 2.6M audio-only dataset, referring to this configuration as the \emph{Baseline}. We then evaluate two audio feature compression methods: 
(i) Applying 1-D convolution with kernel size 3 and stride 2 before audio projector, or 
(ii) Applying average or max pooling with kernel size 2 before audio projector.
We assess performance on audio understanding benchmarks, including Librispeech, Gigaspeech, VoxPopuli, and Long Audio Bench~\citep{goel2025audio} and present results in Table~\ref{tab:downsampling_ablation}.
We also report the embedding per minute of input audio and the average end-to-end latency of the LLM forward pass on Long Audio Bench for each variant in the table.

\begin{table}[h]
  \centering
  \caption{Downsampling method comparison for audio token compression in \modelname. 
  For Librispeech, Gigaspeech, and VoxPopuli we report WER (lower is better). 
  For Long Audio Bench we report accuracy (higher is better) and latency (lower is better).
  Gains are computed relative to the baseline (All audio tokens).}
  \tablestyle{4pt}{1.2}
    \vspace{-10pt}
  \resizebox{.99\linewidth}{!}{
  \begin{tabular}{l | c|ccccc|cc}
    \toprule
     \multicolumn{1}{c|}{\multirow{2}{*}{\textbf{Model}}} & \textbf{Emb./min} & \textbf{Librispeech-cl.} & \textbf{Librispeech-oth.} & \textbf{Gigaspeech} & \textbf{VoxPopuli-ASR} & \multicolumn{2}{c}{\textbf{Long Audio}} \\
     & ($\downarrow$) & WER ($\downarrow$) &WER  ($\downarrow$) & WER ($\downarrow$) & WER ($\downarrow$) & Acc. ($\uparrow$) & Lat. ($\downarrow$) \\
    \midrule
     Baseline - All audio tokens & 750 & 1.91 & 4.49 & 10.77 & 5.89 & 41.28 & 1.78 \\
     \hline
    \textbf{Audio Compression} & - & - & - & - & - & - & - \\
    \ \ Conv1D stride 2 & 375 
      & 2.10\textsubscript{\textcolor{deepgreen}{-0.19}} 
      & 5.22\textsubscript{\textcolor{deepgreen}{-0.73}} 
      & 11.01\textsubscript{\textcolor{deepgreen}{-0.24}} 
      & 6.25\textsubscript{\textcolor{deepgreen}{-0.36}} 
      & 41.79\textsubscript{\textcolor{deepgreen}{+0.51}} 
      & 1.45\textsubscript{\textcolor{deepgreen}{+0.33}} \\
    \ \ Avg. pooling & 375 
      & \underline{1.96}\textsubscript{\textcolor{deepgreen}{-0.05}} 
      & \textbf{4.75}\textsubscript{\textcolor{deepgreen}{-0.26}} 
      & \underline{10.85}\textsubscript{\textcolor{deepgreen}{-0.08}} 
      & \underline{6.24}\textsubscript{\textcolor{deepgreen}{-0.35}} 
      & \underline{42.16}\textsubscript{\textcolor{deepgreen}{+0.88}} 
      & 1.41\textsubscript{\textcolor{deepgreen}{+0.37}} \\
    \rowcolor{lightgreen}
    \ \ Max pooling -- \textbf{chosen} & 375 
      & \textbf{1.93}\textsubscript{\textcolor{deepgreen}{-0.02}} 
      & \underline{4.99}\textsubscript{\textcolor{deepgreen}{-0.50}} 
      & \textbf{10.78}\textsubscript{\textcolor{deepgreen}{-0.01}} 
      & \textbf{6.17}\textsubscript{\textcolor{deepgreen}{-0.28}} 
      & \textbf{43.15}\textsubscript{\textcolor{deepgreen}{+1.87}} 
      & \textbf{1.40}\textsubscript{\textcolor{deepgreen}{+0.38}} \\
    \bottomrule
  \end{tabular}
  }
  \label{tab:downsampling_ablation}
\end{table}

We observe several advantages via compression.
Halving audio tokens leads to significantly shorter latency, from 1.78 sec/sample to 1.40 sec/sample (+17.7\% improvement). For the long audio understanding task, applying audio token downsampling improves the accuracy by 2\% as it compresses information into a more condense representative embeddings, alleviates the burden on LLMs when handling large volumes of audio embeddings.
For short-form benchmarks, we study varying downsampling options, where 
we observe max pooling maintains performance across benchmarks without minimal accuracy degradations. 

\subsection{Model Quantization and Efficient Deployment}
\label{sec:quant}

\begin{figure*}[t]
\centering
\includegraphics[width=1\linewidth]{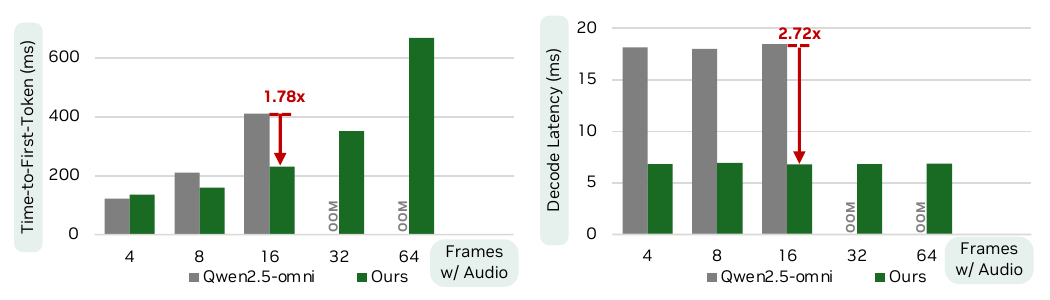}
    \vspace{-10pt}
\caption{Latency comparison between Qwen2.5-Omni and our \modelname model on a GeForce RTX 4090 GPU. Our model achieves 1.7$\times$ faster time-to-first-token latency and 2.72$\times$ faster decoding latency.
}
\label{fig:profile_4090}
\end{figure*}

Although \modelname demonstrates strong omni-modal performance,  real-world deployment quickly encounters multiple constraints. 
Large models or long video sequences often exceed device memory capacity, 
while interactive applications demand extremely low latency. 
To meet these challenges, we compress the model via quantization 
and optimize the system for speedup. 
A detailed analysis of the inference pipeline reveals distinct bottlenecks: 
the vision and audio towers are dominated by dense matrix multiplications, 
processing large batches of tokens in parallel and thus primarily computation-bound; 
in contrast, the LLM decoding stage—where the model consumes and generates one token at a time— 
is memory-bandwidth limited and becomes the key latency bottleneck in long-context scenarios. 
To address this, we adopt a component-aware quantization strategy. 
For the vision and audio towers, we apply {W8A8} quantization, 
reducing arithmetic cost while preserving representational quality. 
For the LLM, we employ {W4A16} quantization, 
compressing weights into 4 bits while retaining 16-bit computation, 
which accelerates bandwidth-limited decoding. 
Finally, to recover accuracy, we integrate 
{Activation-Aware Weight Quantization (AWQ)}~\citep{lin2024awq} 
and {SmoothQuant}~\citep{xiao2023smoothquant}. 

We measure the time-to-first-token latency and decoding latency on a single GeForce RTX 4090 GPU using video clips ranging from 2 to 32 seconds (at 2 frames per second), and compare the performance against Qwen2.5-Omni in Figure~\ref{fig:profile_4090}. Overall, these quantization methods allow a 8B model to handle videos of up to 64 frames on a 24GB RTX 4090 GPU, while achieving 1.7$\times$ lower time-to-first-token latency and 2.72$\times$ faster decoding latency.
For a 16-frame video with audio stream, our model needs only around 160ms to produce the first token.

\subsection{\modelname with ASR Test-Time Scaling Methods}
\label{appendix:e2:asr}
\begin{figure}
    \centering
    \includegraphics[width=0.8\linewidth]{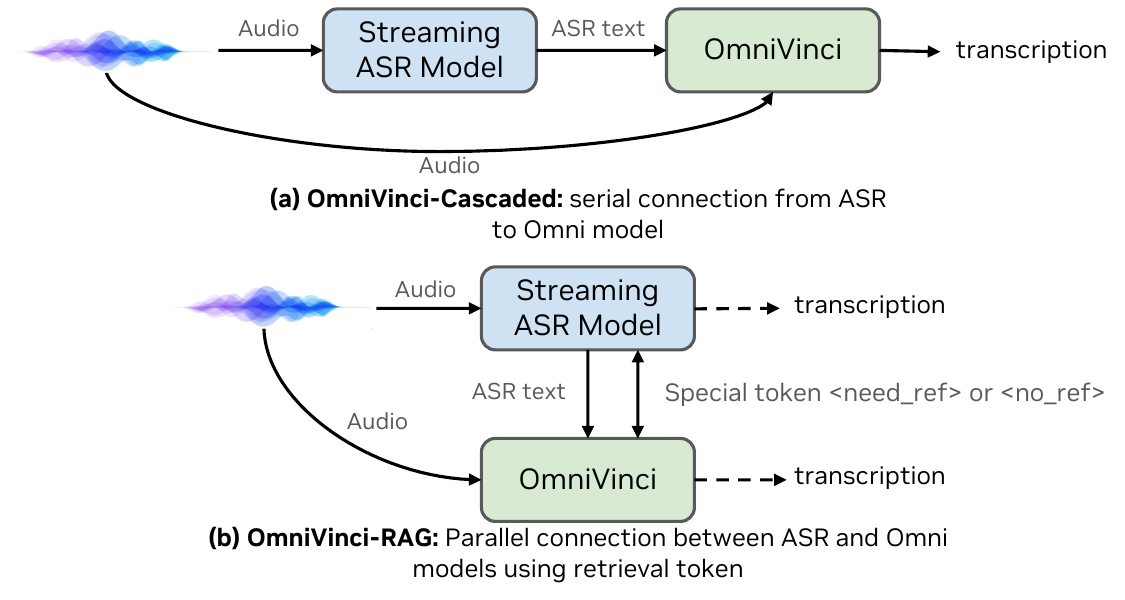}
    \caption{We illustrate two test-time scaling methods using an extra ASR model: (a) \modelname-Cascaded, using ASR history as an additional input to the Omni model with the audio inputs, and (b) \modelname-RAG, using the retrieval token for prediction. The related results are reported in Table~\ref{tab:asr_appendix}.}
    \label{fig:asr:test-time}
\end{figure}

To push the limit of transcription accuracy, we investigate our model's ability to leverage pretrained ASR models in downstream speech understanding tasks. In a cascaded post-ASR processing setup~\citep{yang2023generative} as shown in Figure \ref{fig:asr:test-time} (a), speech inputs are first transcribed by the model’s ASR module and then processed by LLM based generative ASR error correction. We use a popular 800M streaming variant of Whisper-v3-Turbo from SimulStreaming as the cascaded ASR module.

The results are also shown in Table~\ref{tab:asr_appendix}. The cascaded pipeline yields additional improvements on ASR tasks, making it particularly beneficial in offline transcription scenarios. We use Phi-4-mm-instruct 's 5-shot~\citep{abouelenin2025phi} speech modeling setup as one test-time baseline. For Qwen2.5-Omni experiment, we follow the official inference script\footnote{We follow the official ASR cookbook in \url{https://github.com/QwenLM/Qwen2.5-Omni/blob/main/cookbooks/universal_audio_understanding.ipynb} and a related discussion in \url{https://github.com/QwenLM/Qwen2.5-Omni/issues/79} on the Omni settings used in our evaluation.} for the evaluation reported in the fourth row of Table~\ref{tab:asr_appendix}, with the original results shown in the third row. From the extended test-time scaling results, \modelname -cascaded improves average WER from 6.3 to 5.7. The \modelname-RAG setup yields a further improvement, reducing average WER from 6.3 to 5.0 with the same model size of ASR parallel cascading by using ASR text as index for \modelname on mutlimodal ASR correction~\citep{lin2025neko}. We introduce the retriever training details of this setup in the following section.

\begin{table}[ht!]
  \centering
  \caption{Speech Recognition WER~(\%) comparison of different models on speech recognition datasets. }
  \tablestyle{4.2pt}{1.2}
  \scriptsize
  \vspace{-10pt}
  \begin{tabular}{lccccccc}
    \toprule
    \multicolumn{1}{c}{\multirow{2}{*}{\textbf{Model}}} 
      & \multicolumn{6}{c}{\textbf{WER ($\downarrow$)}} \\
    & LS$_{\text{clean}}$ & LS$_{\text{other}}$ 
      & {AMI} & {Tedlium} & {Voxpopuli} & {Avg.}  \\
    \midrule
    Phi-4-MM             & 1.7 & 3.8 & \textbf{11.5} & \textbf{2.9} & 5.9 & 5.2  \\
     Phi-4-MM-in-context (5-shots)             & 1.6 & 3.6 & \textbf{11.5} & 3.0 & 6.1 & 5.2  \\
    \midrule
    Qwen2.5-omni: reported~\citep{xu2025qwen25omni}             & 1.8 & 3.4 & - & - & 5.8 & -  \\
    Qwen2.5-omni: reproduced           & 2.1 & 3.8 & 17.8 & 5.2 & 6.1 & 7.0  \\
    \midrule
    \textbf{\modelname}         & 1.7 & 3.7 & 16.1 & 3.4 & 6.8 &  6.3 \\
    \textbf{\modelname}-cascaded         & 1.6 & \textbf{3.0} & 14.1 & 3.3 & 6.5 &  {5.7} \\
    \textbf{\modelname}-RAG        & \textbf{1.5} & \textbf{3.0} & 11.6 & 3.0 & \textbf{5.7} &  \textbf{5.0} \\
    \bottomrule
  \end{tabular}
  \label{tab:asr_appendix}
\end{table}

\noindent \textbf{\modelname with ASR based Retriever-Augmented Training.}
\label{sec:rag}

As shown in Figure \ref{fig:asr:test-time} (b), given a primary acoustic input, $\mathcal{A}$, our objective is to generate a final, high-fidelity textual output $\mathcal{T}_{\text{final}}$ (either a transcription for ASR or a translation for ST). The model has access to two streams of textual information:
\begin{enumerate}
    \item \textbf{Internal Hypothesis ($\mathcal{T}_{\text{internal}}$):} A first-pass generation produced by the omni-modal model itself, conditioned solely on the acoustic input $\mathcal{A}$. This represents the model's direct, audio-grounded interpretation.
    \item \textbf{External References ($\mathcal{H}$):} A set of candidate transcriptions, $\mathcal{H} = \{h_1, h_2, \dots, h_N\}$, generated by one or more external systems. This set represents external, text-only evidence that may contain valuable corrections or introduce noise.
\end{enumerate}

The task is formulated as a conditional generation problem that jointly models the final output and a decision variable, $d$. The model learns to generate a control token indicating its strategy, followed by the refined text. This decision hinges on the model's ability to align $\mathcal{A}$, $\mathcal{T}_{\text{internal}}$, and $\mathcal{H}$ to determine the most reliable path to the ground truth, $\mathcal{T}_{\text{gt}}$.

\subsubsection{Instructional Formatting for Cross-Modal Decision Making}

To facilitate this decision-making process, we structure the input as a comprehensive instruction that forces the model to weigh different sources of evidence. The model is presented with all modalities and explicitly prompted to declare its generation strategy.

\begin{quote}
\small
\texttt{\textbf{Task:} Perform reference-augmented correction for a given speech input.}\\
\texttt{\textbf{Objective:} Evaluate the quality of an internally generated hypothesis against external candidates. First, select a generation strategy by producing a control token. Then, generate the final, corrected text.}\\
\texttt{- If the internal hypothesis is deemed superior and well-aligned with the audio, select <accept\_internal>.}\\
\texttt{- If the external candidates provide necessary corrections, select <integrate\_reference>.}\\
\texttt{---}\\
\texttt{\textbf{Acoustic Evidence:} [AUDIO]}\\
\texttt{\textbf{External Candidate Transcriptions:}}\\
\texttt{1. \{h\_1\}}\\
\texttt{2. \{h\_2\}}\\
\texttt{3. \{h\_3\}}\\
\texttt{4. \{h\_4\}}\\
\texttt{5. \{h\_5\}}\\
\texttt{\textbf{Internal Hypothesis:}}\\
\texttt{\{T\_internal\}}\\
\texttt{---}\\
\texttt{\textbf{Output:}}
\end{quote}

The model is then trained to generate the complete target string, beginning with either \texttt{<accept\_internal>} or \texttt{<integrate\_reference>}, followed by the corrected and finalized text. We expand the model's vocabulary with these two special tokens to serve as explicit control signals.

\subsubsection{Supervision for Decision-Aware Fine-Tuning}

Supervision for this decision-aware fine-tuning is derived by comparing the internal hypothesis ($\mathcal{T}_{\text{internal}}$) against the ground truth ($\mathcal{T}_{\text{gt}}$) and the external references ($\mathcal{H}$). The decision label is determined as follows:
\begin{itemize}
    \item \textbf{\texttt{<accept\_internal>}:} This label is assigned when the word error rate (WER) of $\mathcal{T}_{\text{internal}}$ is below a predefined threshold or when the external references in $\mathcal{H}$ offer no improvement or introduce hallucinations. This teaches the model to trust its own cross-modal alignment between audio and text when its confidence is high.
    \item \textbf{\texttt{<integrate\_reference>}:} This label is assigned when $\mathcal{T}_{\text{internal}}$ contains correctable errors and at least one hypothesis in $\mathcal{H}$ provides information that reduces the WER relative to $\mathcal{T}_{\text{gt}}$. This trains the model to identify valuable external information and integrate it, effectively re-aligning its understanding based on supplementary textual evidence.
\end{itemize}
The final training target is the concatenation of the assigned decision token and the ground-truth transcript $\mathcal{T}_{\text{gt}}$.

\subsubsection{Inference-Time Control Flow}

At inference, the omni-modal model processes the multi-source input containing the audio, its internal hypothesis, and the external references. The first token generated by the model dictates the subsequent control flow:
\begin{itemize}
    \item If the model generates \texttt{<accept\_internal>}, it signals high confidence in its own audio-to-text mapping. For the final output, we can simply use its pre-computed internal hypothesis, $\mathcal{T}_{\text{internal}}$, or allow the model to regenerate it.
    \item If the model generates \texttt{<integrate\_reference>}, it indicates that the external textual evidence is necessary for achieving a better output. The full sequence generated by the model following this token is taken as the final, corrected transcript.
\end{itemize}
This mechanism provides an interpretable and controllable framework for test-time adaptation, allowing the model to dynamically adjust its reliance on external knowledge based on the specific challenges of each input. This is critical for robust performance in both ASR, where the focus is on transcription fidelity, and ST, where a correct semantic understanding grounded in both audio and reference text is paramount for accurate translation.
Table~\ref{tab:asr_appendix} presents the performance of this method, denoted as \modelname-RAG. It substantially improves the model’s results across all speech recognition benchmarks.

\subsection{Speech Output}
\label{sec:exp_speech_output}

Rather than training a speech generation model from the ground up, we leverage state-of-the-art pre-trained text-to-speech (TTS) systems to produce speech in relevant scenarios, and adapt our approach using a speech codec when needed. Our evaluation focuses on English omni-modal-in and voice-out, using two complementary metrics: mean opinion score (MOS; higher indicates greater naturalness) and TTS word error rate (WER; lower indicates higher intelligibility), the latter measured through an external ASR system. As reported in Table~\ref{tab:mos-wer}, existing off-the-shelf models already yield high-quality, neutral speech suitable for assistant-style applications. Among the back ends tested, \modelname-Magpie achieves the best overall balance (MOS \textbf{4.63}, WER \textbf{2.7}\%), followed closely by gpt-4o-mini-tts (MOS 4.59, WER 3.1\%) and Qwen-omni (MOS 4.53, WER 3.2\%). \modelname-StableCodec delivers a competitive WER (2.9\%) but with slightly reduced naturalness (MOS 4.12), highlighting that intelligibility and perceived naturalness are not always aligned. In contrast, Bark underperforms on both measures (MOS 3.32, WER 8.2\%), consistent with its more stochastic synthesis approach.

\noindent\textbf{Setup.}
We evaluate prompt following on VoiceBench style/control splits and conversational control tasks. We compare three prompting strategies over interleaved audio–vision contexts:
(i) \emph{Transcript prompting} (ASR$\to$text): $[\text{aud},\text{vis}]^{\times 3} + \text{text-prompt}$,
(ii) \emph{Native audio prompting} (encoder features): $[\text{aud},\text{vis}]^{\times 3} + \text{aud-prompt}$,
(iii) \emph{TTS-injected prompting} (render text to speech, then encode): $[\text{aud},\text{vis}]^{\times 3} + \text{TTS}(\text{text-prompt})$.
We also ablate prompt position: \emph{prefix} $[\text{aud-prompt}] + [\text{aud},\text{vis}]^{\times 3}$, \emph{mid} $[\text{aud},\text{vis}], [\text{aud-prompt}], [\text{aud},\text{vis}]^{\times 2}$, and \emph{suffix} $[\text{aud},\text{vis}]^{\times 3}, [\text{aud-prompt}]$.

\noindent\textbf{Metrics.}
We report (a) \emph{Prompt Adherence Rate} (PAR; judged by paired preference and rubric scoring), (b) \emph{slot accuracy} for constrained commands (names, numerals, entities), and (c) latency/efficiency (no additional ASR pass). For speech rendering quality, MOS/WER results are summarized in Table~\ref{tab:mos-wer}.

\begin{tcolorbox}[highlightstyle]
\keyinsight{} 
(1) \emph{Native audio prompting} is the most robust to accents, background noise, and overlapped speech; it preserves prosodic cues (rate, emphasis) that pure transcripts discard, leading to higher PAR and slot accuracy in noisy and accented conditions.
(2) \emph{Transcript prompting} is competitive on clean speech but degrades when ASR struggles on named entities or code-switched fragments.
(3) \emph{TTS-injected prompting} reduces acoustic mismatch in far-field scenarios and is effective when a consistent house voice is desired, but it transfers less speaker/style information than using the raw prompt audio.
(4) Prompt \emph{suffix} placement—immediately before the model’s response—consistently outperforms prefix and mid insertion, likely due to reduced long-range interference in the attention context.
\end{tcolorbox}

Encoding the \emph{audio} prompt directly (no external ASR) yields the best prompt following under realistic noise/accents while lowering latency and memory by avoiding an extra ASR pass. Suffix-position audio prompts provide the strongest control.

\begin{table}[h]
  \centering
  \caption{English naturalness MOS (higher is better) and TTS word error rate (WER; lower is better). Best per column in \textbf{bold}. 
  }
  \label{tab:mos-wer}
    \scriptsize
  \begin{tabular}{@{} l l c c  @{}} %
    \toprule
    \multicolumn{1}{c}{Setup} & \multicolumn{1}{c}{Regime} & \multicolumn{1}{c}{MOS $\uparrow$} & \multicolumn{1}{c}{WER (\%) $\downarrow$} \\
    \midrule
    Qwen-Omni          & auto-regressive & 4.53          & 3.2          \\
    GPT-4o-mini    & --         & 4.59          & 3.1          \\
    \midrule
    \modelname-CozyVoice   & agentic cascaded        & 4.54          & 3.0          \\
    \modelname-Bark        & agentic cascaded        & 3.32          & 8.2          \\
    \modelname-StableCodec & auto-regressive & 4.12          & 2.9          \\
    \modelname-Magpie (\textbf{chosen})      & agentic cascaded         & \textbf{4.63} & \textbf{2.7} \\
    \bottomrule
  \end{tabular}
\end{table}

Beyond raw scores, we observe consistent performance across synthesis {regimes}. Agentic cascaded setups that decouple text planning from acoustic rendering tend to produce strong MOS and low WER in our pipeline, while auto-regressive models are competitive but show greater variance. Importantly, swapping the TTS back end does not alter \modelname’s language understanding or response planning; it only affects the surface realization of speech, simplifying deployment-time customization (\textit{e.g.}, voice, rate).

For interactive agents, streaming synthesis and low perceived latency are crucial. Our chosen back ends support incremental generation, enabling prompt first-audio while the remainder of the utterance is synthesized. In production, we prioritize (i) stability on numerals, abbreviations, and named entities, (ii) speaker consistency across turns, and (iii) graceful handling of punctuation and prosody cues from text.

\end{document}